\documentclass[10pt,journal,compsoc]{IEEEtran}
\usepackage[utf8]{inputenc}
\usepackage{times}
\usepackage{textcomp}
\usepackage{stfloats}
\usepackage{ragged2e}
\usepackage{url}
\usepackage{verbatim}
\usepackage{graphicx}
\usepackage[marginal]{footmisc}
\usepackage{epsfig}
\usepackage{amssymb}
\usepackage{multirow}
\usepackage{multicol}
\usepackage{booktabs}
\graphicspath{{figures/}}
\usepackage{ragged2e}
\usepackage{amsmath}
\usepackage[breaklinks,colorlinks,citecolor=blue,linkcolor=blue,urlcolor=blue,hypertexnames=false]{hyperref}
\usepackage{cleveref}
\usepackage{marvosym}
\usepackage{cite}
\usepackage{tabularx}
\usepackage{makecell}
\usepackage{rotating} 
\usepackage{pifont}
\usepackage{subfigure}
\usepackage{comment}
\usepackage{caption}
\usepackage{longtable}
\usepackage{sidecap}
\usepackage{pifont}
\bstctlcite{IEEEexample:BSTcontrol}

\usepackage{color}
\usepackage{colortbl}%

\usepackage{amsmath,amsfonts}
\usepackage{array}
\def\BibTeX{{\rm B\kern-.05em{\sc i\kern-.025em b}\kern-.08em
    T\kern-.1667em\lower.7ex\hbox{E}\kern-.125emX}}
\usepackage{balance}
\usepackage{booktabs,multirow,tabularx,makecell,adjustbox}
\usepackage{enumitem}

\usepackage{algorithm,algpseudocode}
\usepackage{xcolor}
\usepackage{tikz}
\usetikzlibrary{fit,calc}

\usepackage{xr}
\makeatletter

\newcounter{algsubstate}

\colorlet{pink}{red!40}
\colorlet{blue2}{cyan!60}
\colorlet{green}{green!40}

\usepackage{amsmath}
\usepackage{booktabs}
\usepackage{graphicx}

\usepackage{threeparttable}
\usepackage{colortbl}
\usepackage{multirow}
\usepackage{graphicx}
\usepackage{array}

\usepackage{graphicx} 
\usepackage{textcomp}

\usepackage{threeparttable}
\usepackage{colortbl}
\usepackage{multirow}
\usepackage{graphicx}
\usepackage{array}

\usepackage{ragged2e}  
\usepackage[edges]{forest}
\usepackage{tikz}
\usepackage{verbatim}

\usepackage{eso-pic}   
\usepackage{graphicx}  

\definecolor{mycolor_0}{RGB}{150, 200, 180}
\definecolor{mycolor_1}{RGB}{245, 239, 200}  
\definecolor{mycolor_2}{RGB}{202, 232, 252}  
\definecolor{mycolor_3}{RGB}{214, 238, 214}  
\definecolor{mycolor_4}{RGB}{233, 218, 248}  
\definecolor{mycolor_5}{RGB}{255, 230, 204}  
\definecolor{mycolor_6}{RGB}{204, 255, 255}  
\definecolor{mycolor_7}{RGB}{255, 204, 229}  
\definecolor{mycolor_8}{RGB}{230, 230, 250}  
\definecolor{mygreen}{rgb}{0.0, 0.5, 0.0}  
\usepackage{xcolor}

\tikzstyle{leaf}=[draw=hiddendraw,
    rounded corners, minimum height=1em,
    fill=mygreen!40,text opacity=1, 
    fill opacity=.5,  text=black,align=left,font=\scriptsize,
    inner xsep=3pt,
    inner ysep=0.3pt,
    ]
\tikzstyle{middle}=[draw=hiddendraw,
    rounded corners, minimum height=1em,
    fill=output-white!40,text opacity=1, 
    fill opacity=.5,  text=black,align=center,font=\scriptsize,
    inner xsep=7pt,
    inner ysep=0.3pt,
    ]
    
\newcolumntype{C}[1]{>{\centering\arraybackslash}p{#1}}

\begin{document}

\title{\includegraphics[height=1.5ex]{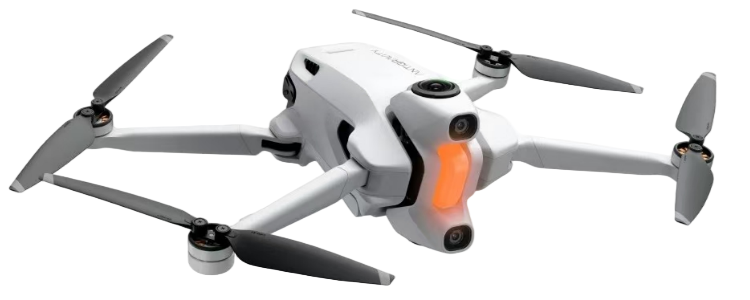}One Flight Over the Gap: A Survey from Perspective to Panoramic Vision}

\author{Xin Lin,
Xian Ge,
Dizhe Zhang,
Zhaoliang Wan,
Xianshun Wang,
Xiangtai Li, \\
Wenjie Jiang,
Bo Du,
Dacheng Tao,
Ming-Hsuan Yang,
Lu Qi
\IEEEcompsocitemizethanks{
\IEEEcompsocthanksitem Xin Lin, Xian Ge, Dizhe Zhang, Zhaoliang Wan, Xianshun Wang, Wenjie Jiang, and Lu Qi are with Insta360 Research. 
Xin Lin and Ming-Hsuan Yang are with the University of California, San Diego and the University of California, Merced. 
Xiangtai Li and Dachenng Tao are with Nanyang Technological University. 
Lu Qi and Bo Du are with Wuhan University. 
Work was done during Xin Lin’s internship at Insta360 Research. 
Xin Lin and Xian Ge share equal contributions. Dizhe Zhang is the project leader. 
Dizhe Zhang and Lu Qi are the corresponding authors.}
}

\IEEEtitleabstractindextext{
\begin{abstract}
\justifying
Driven by the demand for spatial intelligence and holistic scene perception, omnidirectional images (ODIs), which provide a complete 360\textdegree{} field of view, are receiving growing attention across diverse applications such as virtual reality, autonomous driving, and embodied robotics. Despite their unique characteristics, ODIs exhibit remarkable differences from perspective images in terms of geometric projection, spatial distribution, and boundary continuity, making it challenging for direct domain adaption from existing perspective-based methods. In this survey, we present a comprehensive review of recent techniques in panoramic vision with a particular emphasis on the perspective-to-panorama adaptation problem. At first, we revisit the panoramic imaging pipeline and projection methods to build the prior knowledge required for analyzing the structural disparities between ODIs and perspective ones. Then, we summarize three challenges of domain adaptation, including severe geometric distortions near the poles, the non-uniform sampling in Equirectangular Projection (ERP), and the periodic continuity of panoramic boundaries. Based on the discussions above, we cover 20+ representative tasks drawn from more than 300 research papers in two dimensions. On one hand, we present a cross-method analysis of representative strategies for addressing panoramic specific challenges across different tasks. On the other hand, we conduct a cross-task comparison and classify panoramic vision into four major categories: visual quality enhancement and assessment, visual understanding, multimodal understanding, and visual generation. In addition, we discuss open challenges and future directions, emphasizing data, models, and applications that will drive the advancement of panoramic vision research. Compared to previous surveys that focused on task-specific pipelines, ours has a more unified and evolving landscape of panoramic visual learning. We hope that our work can provide new insight and forward-looking perspectives to advance the development of panoramic vision technologies. Our project page is \url{https://insta360-research-team.github.io/Survey-of-Panorama/}.
\end{abstract}

}

\maketitle
\IEEEdisplaynontitleabstractindextext
\IEEEpeerreviewmaketitle

\IEEEraisesectionheading{\section{Introduction}}

\begin{figure*}[!h]
  \centering
    \vspace{-1.0cm}
  \includegraphics[width=1\linewidth]{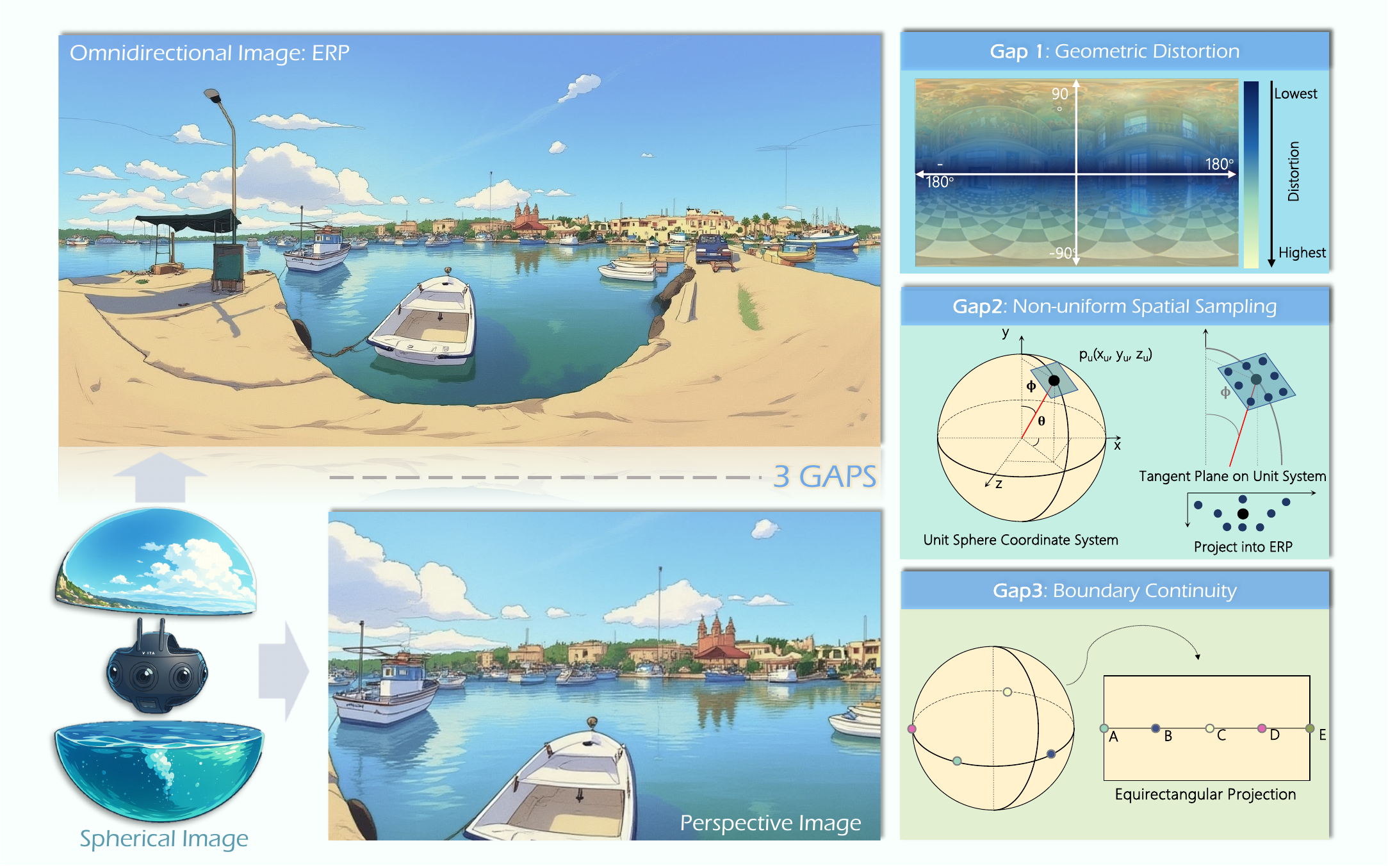}
  \vspace{-15pt}
  \caption{From spherical image to Panoramic ERP and perspective image, ERP preserves a complete field of view compared to perspective images, but introduces three major domain gaps: (1) geometric distortion, (2) non-uniform spatial sampling, and (3) boundary continuity.}
  \label{fig:teaser}
  \vspace{-15pt}
\end{figure*}

In recent years, computer vision techniques have made significant progress in understanding 2D perspective images, benefiting a wide range of tasks, including recognition, reconstruction, and generation, in numerous real-world applications. 
Driven by deep learning, many classic architectures and learning paradigms have been developed under the camera's assumptions of perspective projections, supported by publicly available datasets~\cite{deng2009imagenet,lin2014microsoft,cordts2016cityscapes,qi2022high} and widespread real-world deployment~\cite{qin2024restore,ren2024ninth}.
However, with the growing demand for immersive perception and holistic scene understanding, omnidirectional images (ODIs), which provide a complete 360\textdegree{} field of view, have drawn increasing attention from the research community.
Compared to conventional perspective images, ODIs can provide broader spatial coverage and richer contextual information, making them indispensable for emerging applications such as virtual reality (VR)~\cite{linhqgs}, autonomous driving~\cite{qi2019amodal}, and embodied robotics~\cite{wan2025rapid}.

Despite their potential, ODIs differ significantly from perspective images in terms of imaging geometry.
As illustrated on the right side of Fig.~\ref{fig:teaser}, panoramic representations introduce unique challenges, including geometric distortion, uneven spatial sampling, and boundary continuities, which are especially common in standard formats such as Equirectangular Projection (ERP).
These differences result in an extreme domain gap, where methods trained in perspective images often fail to generalize effectively to panoramic scenarios. 
The planar assumptions embedded in conventional deep models hinder their ability to handle spherical geometry and full-scene coverage, thus limiting the adaptability of perspective-based techniques and slowing progress in omnidirectional vision. 
Then, methods specifically designed for panoramic vision have emerged. 
Unless otherwise specified, we use the terms omnidirectional and panoramic to represent a 360\textdegree{} view, as both terms are widely used.

During the past decade, several surveys have reviewed specific aspects of omnidirectional vision, including 360\textdegree{} video streaming and compression~\cite{zink2019scalable, yaqoob2020survey}, visual quality assessment~\cite{sui2024survey}, indoor layout estimation~\cite{zou2019manhattan}, super-resolution~\cite{zhao2025systematic}, optical systems~\cite{gao2022review}, and 3D perception tasks~\cite{jiang20243d, lopes20243d, yu2023applications, meng20253d}.
More recently, a review~\cite{ai2025survey} has provided a system-level overview of deep learning applications in panoramic vision.
In contrast to their structural-paradigm-based categorization, our work begins from the more fundamental perspective–panorama gap, thoroughly examining task-specific differences between perspective and panoramic representations, and systematically analyzing the resulting methodological variations. 
We aim to provide methodology-level insights for addressing panoramic vision tasks while integrating promising emerging technologies to broaden future research directions.

By this motivation, we investigate various ODI methods for each specific task from a perspective-to-panorama viewpoint, analyzing the strategies and efforts to bridge the domain gap from both vertical (cross-method) and horizontal (cross-task) perspectives.
Moreover, we place particular emphasis on two aspects.
On one hand, we highlight ODI imaging systems and several emerging and rapidly evolving techniques, emphasizing the potential of diffusion- or auto-regressive- or 3D-reconstruction-based generative paradigms guided by ODI priors. 
On the other hand, the limitations of existing approaches and promising future directions are discussed.
Together, these dimensions bring a holistic understanding of the methodological landscape in omnidirectional vision and uncover opportunities for innovation at the intersection of geometry, semantics, and generation.

To this end, this survey reviews over 20 representative tasks based on over 300 research papers, and is organized into several core sections, each focusing on a key component of panoramic vision.
Section~\ref{sec:is} revisits the panoramic imaging pipeline, from acquisition to stitching and projection, bringing with a clear foundation for understanding panoramic–perspective differences and supports subsequent methodological analysis.
Section~\ref{sec:oc} presents the three intrinsic characteristics of ODIs which distinguish them from perspective images and reveal the roots of the domain gap, followed by a cross-method analysis of representative mitigation strategies.
Section~\ref{sec:pt} conducts a cross-task comparison that synthesizes common insights and highlights methodological trends. It also identifies several rapidly evolving techniques, such as diffusion models, 3D Gaussian Splatting, and multimodal fusion, which are increasingly emerging but remain systematically unexplored in previous surveys.
Last, Section~\ref{sec:future} discusses open challenges and promising future directions, with a focus on data, models, and applications that will advance future research on panoramic vision.

As a unique modality that allows spatially comprehensive 360\textdegree{} perception, panoramic vision demonstrates strong potential and practical value in various applications such as spatial intelligence or immersive interaction. 
Through our comprehensive survey, we identify that bridging existing research gaps by transferring and adapting insights from the conventional perspective-vision domain can substantially benefit omnidirectional computer vision.
We hope that our work can provide more insightful and forward-looking guidance for future research in this field.

\vspace{-0.12in}
\section{Panoramic Imaging Background}
\label{sec:is}
This section presents background knowledge on panoramic imaging, discussing its representative imaging systems, the stitching pipeline, and widely adopted projection formats. Additional details are provided in our supplementary file.
\subsection{Imaging Systems}
\begin{figure*}[t]
    \centering
    \includegraphics[width=1\textwidth]{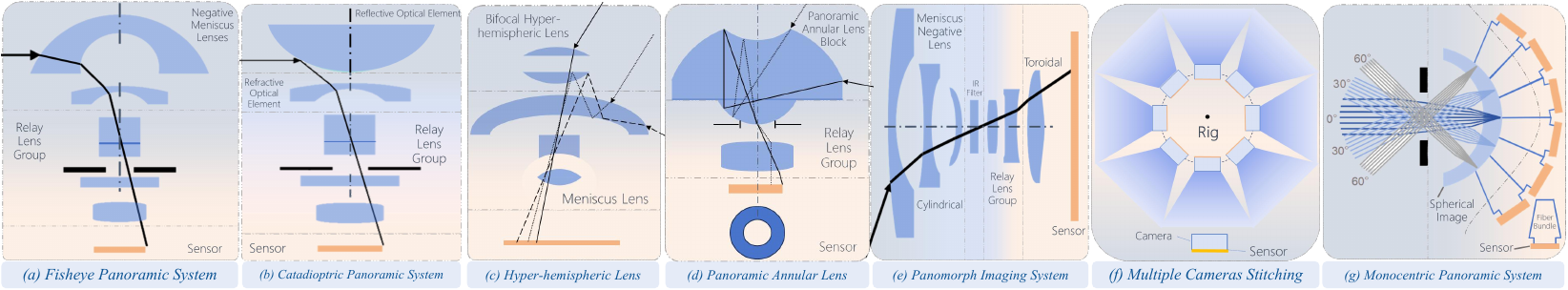}
        \vspace{-15pt}
    \caption{Illustration of Seven Typical Panoramic Imaging Systems: Optical Designs for Capturing 360\textdegree{} Field-of-View.}
    \label{fig:imaging}
    \vspace{-15pt}
\end{figure*}

Panoramic imaging systems capture 360\textdegree{} scenes for holistic perception in vision tasks. Unlike perspective cameras, they achieve ultra-wide fields of view through wide-angle refraction, mirror-based reflection, or multi-camera stitching. In this section, we introduce seven representative designs, as illustrated in Fig.~\ref{fig:imaging}.

\noindent \textbf{Fisheye Panoramic System} is an ultra-wide angle optical system with a field of view (FoV) greater than 180\textdegree{}. 
In Fig.~\ref{fig:imaging}(a), it employs a front group of two to three negative meniscus lenses that compress the wide object-side FoV into a narrower cone, which is then relayed by a subsequent lens group for aberration correction.
Such a system has practical advantages, including compact design, simplified image acquisition, lower manufacturing cost, and improved installation stability. 
Since the optical path is ``folded'' through multiple front stage elements, 
fisheye optics inherently produce substantial distortion, often in the range of 15–20\%, which renders distortion control a central challenge in ultra–wide-angle lens design.

\noindent \textbf{Catadioptric Panoramic System} is shown in Fig.~\ref{fig:imaging}(b), which integrates reflective and refractive optical elements to achieve single-viewpoint 360\textdegree{} capture. 
Incoming light is first redirected by a curved mirror and then focused onto the image sensor through a relay lens group, which also contributes to aberration correction. 
By combining mirror reflection to redirect ambient light into the optical path, catadioptric systems achieve an extended FoV with distortion typically above 10\% while also providing broader coverage and higher spatial resolution compared to fisheye optics.
However, they often suffer from a central blind spot caused by an occlusion from the mirror structure itself.

\noindent \textbf{Hyper-hemispheric Lens System} is illustrated in Fig.~\ref{fig:imaging}(c), which employs a series of meniscus lenses to capture light from the entire 360\textdegree{} surroundings. 
To address the issue of the central blind spot, a forward-view lens group is combined with a mirror to image the blind region, forming two separate optical paths with different focal lengths that include the panoramic and forward channels.
By integrating dual optical paths, hyper-hemispheric systems extend the FoV to 360\textdegree{} $\times$ 260\textdegree{}, providing nearly complete spherical coverage and comprehensive ultra–wide-angle imaging. 

\noindent \textbf{Panoramic Annular Lens}, shown in Fig.~\ref{fig:imaging}(d), employs a coaxial catadioptric configuration that replaces the fisheye’s front lens group with a compact mirror–lens assembly. 
The light from the full 360\textdegree{} surroundings undergoes two refractions and two reflections, forming a narrow-angle beam that passes through the aperture stop and relay lens group before reaching the sensor, producing a 2D annular panoramic image based on a planar cylindrical projection.
This design enables full 360\textdegree{} horizontal FoV with a vertical FoV of $\beta$, while significantly reducing system size and complexity compared to fisheye optics. 
However, the small front mirror inevitably introduces a central blind zone corresponding to a half-FoV angle $\alpha$ and limits its vertical completeness.

\noindent \textbf{Panomorph Imaging System} Fig.~\ref{fig:imaging}(e) enhances sensor utilization by applying spatially varying anamorphic magnification, realized through cylindrical or toroidal optics, to concentrate pixel density in user-defined regions of interest (ROIs). 
Compared to the traditional fisheye lenses that uniformly compress angles, Panomorph optics allocate higher resolution to targeted areas while reducing redundancy elsewhere, thus improving data compression, bandwidth efficiency, and semantic scene understanding.

\noindent \textbf{Single Camera Scanning and Multiple Cameras Stitching} Fig.~\ref{fig:imaging}(f) generate panoramas by stitching images captured from different viewpoints. 
They can be divided into two categories. 
The first method utilizes a single static camera that rotates in place to capture the entire field of view, after which the acquired images are stitched together through geometric alignment techniques.
The second employs a fixed multi-camera rig, in which cameras simultaneously capture images from different directions, followed by stitching into a seamless panorama.
While rotating single-camera systems are low-cost and straightforward, their long scanning process prevents real-time or gaze-based imaging. 
By contrast, multi-camera rigs enable real-time construction but require precise synchronization and calibration to avoid misalignment artifacts.

\noindent \textbf{Monocentric Panoramic System} is presented in Fig.~\ref{fig:imaging}(g), which adopts a spherically concentric architecture, where all optical surfaces share a common center of curvature.
This design is inspired by the compound eyes of arthropods, which enable curved image sensors to be directly coupled with multiple apertures or optical fibers, resulting in a compact yet high-quality panoramic imaging configuration.
By ensuring symmetric light entry, monocentric systems reduce optical aberrations and provide uniform imaging performance across ultra-wide fields of view. 
They are particularly well-suited for multi-aperture or fiber-coupled setups, combining compact design with high-quality imaging.

\subsection{Stitching}

\begin{figure*}[t]
    \centering
    \vspace{-5pt}
    \includegraphics[width=\textwidth]{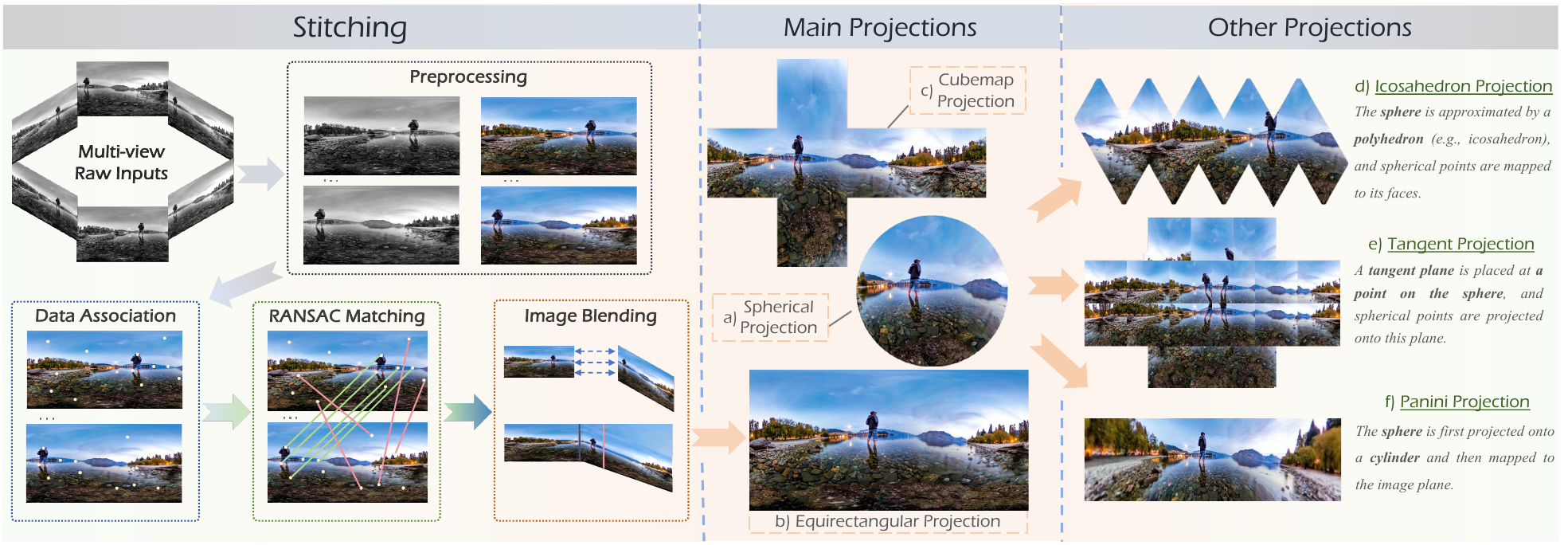}
        \vspace{-15pt}
    \caption{Comprehensive Pipeline for Panorama Stitching: Preprocessing involves classical image signal processing (ISP) steps, including demosaicing, denoising and correction, Data Association through keypoint detection and matching, Geometric Alignment using RANSAC and homography estimation, and Image Blending with gain compensation and straightening. Representative projection methods for 360\textdegree{} images: (a) spherical projection as the foundational representation mapping content onto the unit sphere, from which several planar projections are derived, including (b) equirectangular projection with longitude–latitude mapping, (c) cubemap projection with six 90\textdegree{} perspective faces, (d) icosahedron projection with near-uniform sampling, (e) tangent projection via local planar mapping, and (f) panini projection preserving vertical lines while compressing horizontal fields.}
    \label{fig:projection}
    \vspace{-15pt}
\end{figure*}

Panorama stitching refers to the process of aligning and blending a set of images that cover a 360\textdegree{} view into a seamless panoramic image. 
In the left side of Fig.~\ref{fig:projection}, we illustrate the typical pipeline, which includes data pre-processing, data association, geometric alignment, and image blending.

\noindent \textbf{Data Preprocessing} involves classical image signal processing (ISP) steps such as demosaicing, noise reduction, camera calibration, distortion correction, and exposure/color compensation.
Noise is reduced using filters, with its intensity depending on camera parameters \cite{faraji2006ccd}, while calibration maps 3D world coordinates to 2D pixels, with different camera types (e.g. pinhole, fisheye) requiring distinct models such as polynomial~\cite{basu1995alternative} or Zurich~\cite{scaramuzza2007omnidirectional}.
The distortion correction also compensates for the radial and tangential deviations, and motion estimation or common-view extraction may be further applied for more reliable stitching.

\noindent \textbf{Data Association} establishes alignment across views, typically categorized into three strategies. 
First, spatial association matches current and previous frames using visual features via descriptors such as SIFT~\cite{lowe2004distinctive}, SURF~\cite{bay2006surf}, ORB~\cite{rublee2011orb}, or LSD~\cite{von2012lsd}.
Second, geometric association takes advantage of epipolar constraints to reduce the search space. 
Finally, photometric association utilizes optical flow to extract motion-consistent feature points between frames for temporal correspondence estimation.

\noindent \textbf{Geometric Alignment} ensures robustness and consistency under wide FoV and nonlinear distortions, where planar homographies are often inadequate. 
Outlier rejection (e.g., RANSAC~\cite{fischler1981random}) is essential for reliable transformation estimation, while local warping (e.g., mesh-based) addresses depth variation and parallax. 
Seam optimization techniques, such as graph cut or energy-minimizing seams, are then applied to refine transitions in overlapping regions.

\noindent \textbf{Image Blending} is the final step to ensure seamless visual transitions, compensating for color and lighting variations between images. 
Three mainstream strategies are widely used: (1) linear blending (feathering) achieves smooth transitions through weighted averaging in overlaps; 
(2) multiband blending (Laplacian pyramids) for scale-adaptive fusion and reduced ghosting; 
and (3) Poisson-based fusion for global gradient consistency, with efficient variants such as MVC blending to improve speed.

\subsection{Projection Formats}
Projection is defined as the process of mapping spherical content to 2D formats that serves as the foundation of panoramic vision. 
Different projection schemes aim to balance distortion, directional continuity, and computational efficiency for specific tasks. 
As illustrated in the right-hand side of Fig.~\ref{fig:projection}, there are three main and several other projections.
On one hand, the spherical projection directly represents directions on the unit sphere, the Equirectangular Projection (ERP) is most widely adopted for its simple bijective mapping, and the Cubemap Projection (CMP) alleviates ERP’s severe polar distortion by sampling along cube faces.
On the other hand, more advanced designs such as Icosahedron Projection, Tangent Projection, and Polyhedron Projection further improve geometric fidelity and facilitate compatibility with perspective-based vision models.
In the following, we briefly introduce those six kinds of projections due to the page limitations.
A more detailed description of the projection transformations is provided in the supplementary material.

\noindent \textbf{Spherical Projection.} A 360\textdegree{} camera can be modeled as projecting all visible 3D points onto the surface of a unit sphere. This representation provides a unified, distortion-free view of all directions and serves as the foundation for panoramic imaging.

\noindent \textbf{Equirectangular Projection (ERP).} As the most widely used format, ERP directly unwraps spherical longitude and latitude onto a 2D plane, similar to a world map. While efficient for storage and rendering, it introduces severe distortions near the poles.

\noindent \textbf{Cubemap Projection (CMP).} CMP maps the sphere onto six cube faces, each covering a 90\textdegree{} FoV. This reduces polar distortion compared to ERP and is therefore well-suited for panoramic rendering and processing.

\noindent \textbf{Polyhedron Projection (PP).} PP approximates the sphere with a polyhedron (e.g., icosahedron) and maps spherical points onto its polygonal faces. Recursive subdivision of faces yields nearly uniform sampling with reduced distortion, but inevitably increases overall representation complexity.

\noindent \textbf{Tangent Projection (TP).} TP projects spherical content onto multiple tangent planes placed around the sphere, producing locally distortion-free patches. This enables the reuse of perspective vision models but requires precise stitching across patches.

\noindent \textbf{Panini Projection.} Panini projection reduces distortions of wide-angle rectilinear views ($>70^\circ$) by preserving vertical and radial lines while compressing the horizontal field. It provides a smooth trade-off between central magnification and edge compression.

\section{Structural Challenges and Strategies}
\label{sec:oc}

\begin{figure*}[t]
\centering
\includegraphics[width=1\linewidth]{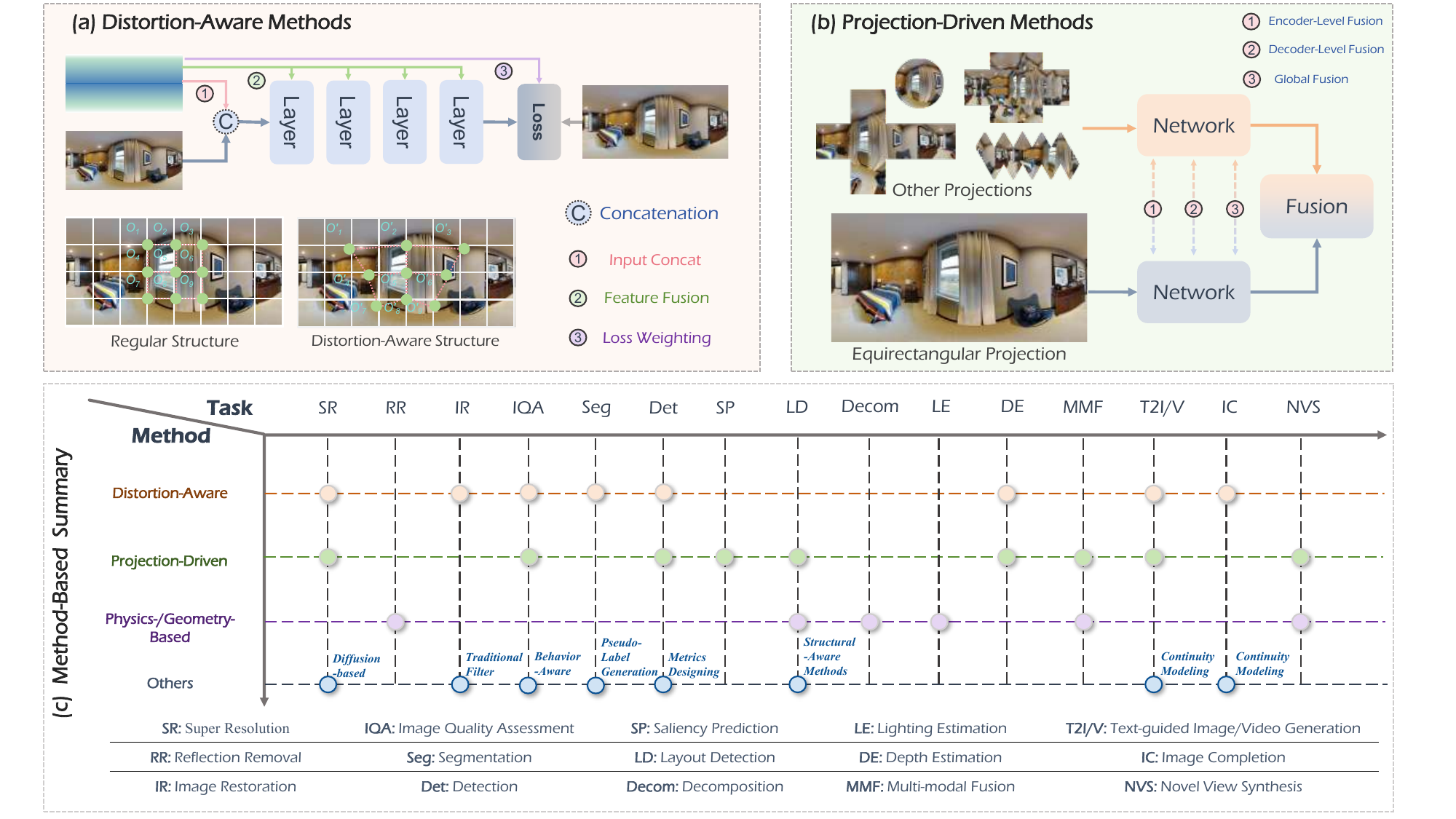}
\caption{Overview of representative strategies for mitigating structural challenges and their task-level summary. 
(a) \textbf{Distortion-Aware Methods} either leverage distortion maps (top) with three representative utilizations-input concatenation, feature fusion, and loss weighting, or design distortion-aware architectures (bottom), such as CNNs, Transformers, and diffusion models with adaptive kernels, attention, or noise initialization. 
(b) \textbf{Projection-Driven Methods} alleviate ERP distortions by re-projecting panoramas into alternative views (e.g., cubemap, tangent plane) and fusing multi-projection features. 
(c) \textbf{Method-Based Summary} highlights the task-level applicability across representative panoramic vision tasks, with distortion-aware and projection-driven methods as two core strategies.
}
\vspace{-15pt}
\label{fig:strategy}
\end{figure*}

Although omnidirectional images (ODI) provide full 360\textdegree{} coverage for immersive perception, their structural differences from perspective images create a domain gap that hinders direct model transfer.
In this section, we analyze three key structural characteristics that distinguish ODIs from perspective images: geometric distortion, non-uniform spatial sampling, and boundary continuity, as shown in the right side of Fig.~\ref{fig:teaser}.
Then, these challenges have motivated a variety of methodological solutions. 
As summarized in Fig.~\ref{fig:strategy}(c), we organize existing approaches through a vertical cross-method analysis into four classes: (1) Distortion-aware, (2) Projection-driven, (3) Physics- or geometry-based, and (4) Other designs, including diffusion, behavior modeling, and metric learning.
Among these, the distortion-aware and projection-driven methods are the most representative and will be summarized in this section.

\subsection{Structural Challenges in ERP-based ODIs}
\label{subsec:challenge}

\noindent \textbf{Geometric Distortion.} 
In ERP, unwrapping the sphere onto a 2D plane introduces distortions that increase with latitude and are most severe near the poles (±90\textdegree{}).
As shown in Fig.~\ref{fig:teaser}, objects near the poles appear significantly stretched and warped, leading to an inaccurate perception of shape and structure. 
Such distortion limits the effectiveness of standard convolutional neural networks (CNNs), whose translation-invariant filters are ill-suited for spherical geometry.
Near the poles, this assumption fails, resulting in a degraded feature extraction.

\noindent \textbf{Non-uniform Spatial Sampling.} Each horizontal line of the ERP corresponds to a constant latitude on the sphere, which leads to the density of pixels to vary across different latitudes. 
As illustrated in Fig.~\ref{fig:teaser}, regions around the equator have dense and fine-grained sampling, whereas the poles exhibit sparse sampling. 
This leads to significant variation in the pixel-to-surface area ratio across the image. 
This is an imbalance that disrupts the scale invariance of visual models, as objects of the same size appear at different resolutions depending on their latitude.

\noindent \textbf{Boundary Continuity.} In contrast to conventional perspective images, ODIs inherently preserve boundary continuity.
Specifically, in ERP, the left and right boundaries correspond to adjacent areas of the sphere, resulting in a seamless horizontal loop.
This property preserves spatial continuity across the image boundaries on the spherical domain.
However, conventional convolutional neural networks (CNNs), even the position embeddings in the Transformer structure, originally designed for perspective images, often treat ERP images as planar and fail to account for this horizontal continuity. 
As a result, visual features that span the spherical seam may be incorrectly treated as disjointed, leading to suboptimal performance near the horizontal boundaries.

\subsection{Method-Aware Comparison}
\label{subsec:method comparison}

Building on the structural challenges discussed in subsection~\ref{subsec:challenge}, we further draw a conclusion of current methods to address the inconsistency between data characteristics and task requirements. 
ODIs suffer from distortion, non-uniform sampling, and boundary continuity, leading different tasks to emphasize either global semantic consistency (e.g., segmentation, restoration) or local geometric precision (e.g., depth, optical flow), which in turn shapes methodological choices.

As summarized in Fig.~\ref{fig:strategy}, existing methodologies can be classified into four categories. 
Distortion-aware methods maintain the ERP format while accounting for distortions.
Projection-driven methods reduce distortion by re-projecting into alternative views. 
Physics- and geometry-based methods integrate priors such as lighting models or spatial layout constraints. 
Finally, other designs cover diffusion-based generative approaches, behavior-aware strategies, and metric-oriented frameworks.
Among these, distortion-aware and projection-driven methods are the two most widely adopted categories, and the following subsections provide a detailed analysis of their strengths, weaknesses, and task-level applicability.

\noindent \textbf{Distortion-Aware Methods} retain the unified ERP representation and embed distortions into the network design, which are shown in Fig. \ref{fig:strategy}(a).
Some works~\cite{osrt, zhao2023stripe, VSBNET, ddcnet, zhang2018saliency, xie2019effective, tateno2018distortion, zhuang2022acdnet} introduce spatially adaptive convolution kernels, where their filter shapes or receptive fields vary with latitude to account for geometric distortion.
Some studies~\cite{osrt, zhao2023stripe, VSBNET, ddcnet, zhang2018saliency, xie2019effective, tateno2018distortion, zhuang2022acdnet} introduce spatially adaptive designs in CNNs, Transformers, and diffusion architectures, where convolution kernels, attention windows, or even the initialization of generative noise are adjusted with latitude to account for geometric distortion and the non-uniform pixel distribution of panoramic images.
Others~\cite{360sisr, sun2023opdn, liu2025distortion} employ distortion maps—precomputed weight masks indicating the severity of distortion at each pixel—to guide feature learning in multiple ways. As illustrated in Fig., distortion maps can be (\ding{172}) concatenated with the input panorama to provide pixel-wise distortion cues, (\ding{173}) fused with intermediate feature layers to modulate representation learning adaptively, and (\ding{174}) incorporated into the loss function as weighted penalties to emphasize errors in highly distorted regions. These strategies collectively compensate for the spatial non-uniformity inherent in ERP images.
Their advantages include: (1) preserving global pixel–semantic correspondence without slicing or projection loss; (2) compatibility with CNN/Transformer/diffusion frameworks for end-to-end training; (3) flexible adaptation via deformable convolutions, re-weighted losses, or distortion maps. Limitations are: (1) residual polar distortion in ERP leading to degraded accuracy in high-deformation regions; (2) reduced robustness in geometry-sensitive tasks (e.g., depth, optical flow, keypoint matching) where precise local geometry is critical.

\noindent \textbf{Projection-Driven Methods} complement ERP with multiple projections that introduce less distortion, thereby alleviating its adverse effects, which are shown in Fig.~\ref{fig:strategy}(b). Representative examples include Cubemap Projection (CP), which reduces polar distortion by splitting the scene into six perspective views, Tangent Projection for locally distortion-free mapping, Polyhedron-based Projections (e.g., icosahedron), and Spherical Projection that directly preserves the underlying geometry. Strengths: (1) effectively suppress distortions, especially at poles and seams; (2) enable direct reuse of perspective models and large pre-trained backbones; (3) achieve stronger performance in geometry-sensitive tasks (e.g., depth, optical flow, NVS); (4) flexibility for task-specific adaptation, as different projections can be selected according to the application. Limitations: (1) fragmented information across projections, requiring additional fusion mechanisms; (2) higher computational and memory overhead from multi-view redundancy; (3) some projections demand bespoke architectures and training.

\noindent \textbf{Applicability Across Tasks.} Task-level analysis shows clear preferences for the two strategies: (1) Distortion-aware methods suit tasks demanding global semantic consistency and perceptual quality, such as super-resolution, restoration, completion, segmentation, and detection. (2) Projection compensation methods excel in geometry-sensitive domains—depth, optical flow, keypoint matching, novel view synthesis (NVS)—and in multi-modal fusion (e.g., LiDAR + panorama, mapping, visual odometry), where alignment with perspective modalities is crucial.
(3) Some tasks admit both strategies depending on application goals. For example, super-resolution can prioritize global consistency and perceptual quality (distortion-aware) for video playback or immersive display, or emphasize local geometric fidelity (projection compensation) for architectural preservation or fine-grained reconstruction. Similarly, text-to-image/video generation benefits from distortion-aware designs for holistic semantic alignment, while projection-based schemes provide finer local control via perspective fusion.
(4) In physics-driven tasks (e.g., reflection removal, parametric lighting estimation, regression-based layout detection), projection choice plays a secondary role compared to physical priors.
(5) For underexplored areas (e.g., tracking, pose estimation, mapping), current evidence is too limited to establish clear preferences, highlighting directions for future study.

\begin{figure*}[t]
\centering
\resizebox{0.9\textwidth}{!}{
\begin{forest}
    for tree={
        forked edges,
        grow=east,
        reversed=true,
        anchor=base west,
        parent anchor=east,
        child anchor=west,
        base=middle,
        font=\tiny,
        rectangle,
        line width=0.2pt,
        draw = black!40,
        rounded corners=2pt,
        align=left,
        minimum width=1.5em,  
        s sep=3pt,            
        l sep=4pt,            
        inner xsep = 0.3pt,     
        inner ysep = 0.3pt, 
        edge path={
            \noexpand\path [draw, \forestoption{edge}]
            (!u.parent anchor) -- ++(1.5mm,0) |- (.child anchor) \forestoption{edge label};},
        ver/.style={rotate=90, child anchor=north, parent anchor=south, anchor=center},
        font=\linespread{1}\selectfont,
    },
    where level=1{font=\tiny,fill=blue!0}{},
    where level=2{font=\tiny,fill=mygreen!5}{}, %
    where level=3{font=\tiny,fill=mygreen!5}{}, %
    where level=4{font=\tiny,fill=mygreen!5}{},
    where level=5{font=\tiny,fill=mygreen!5}{},
    where level=6{font=\tiny,fill=mygreen!5}{},
    [Panorama Tasks, ver, color=mycolor_0, fill=mycolor_0, text=black, font=\small, text width=18em, text centered, inner ysep=4pt
        [Super Resolution, color=mycolor_3, fill=mycolor_3, text=black, inner ysep=0.3pt, text width=8em, text centered
            [{
                \textbf{\ \,Distortion-Aware:} LAUNet~\cite{launet}, 360-SISR~\cite{360sisr}, OSRGAN~\cite{osrgan}, OSRT~\cite{osrt}, An et al. \cite{an2023perception},  OPDN~\cite{sun2023opdn}, FATO~\cite{an2024fato},GDGT-OSR~\cite{gdgtosr}, Cao et al.~\cite{cao2025geometric} \\
                \ \,FAOR~\cite{shen2025fast}, MambaOSR~\cite{wen2025mambaosr}, SMFN~\cite{smfn}, S3PO~\cite{s3po}, STDAN~\cite{an2024spatio}, FOCAS~\cite{wang2023central} \textbf{\ \,Traditional:} SRCNN$\_$FT~\cite{SRCNNFT}, OESRGAN~\cite{oesrgan}
            }, color=mycolor_3, fill=mycolor_3, text=black, inner ysep=0.3pt, text width=36.5em
            ]
            [{
                \textbf{\ \,Projection-Driven:}
                SphereSR~\cite{spheresr}, OmniZoomer~\cite{omnizoomer},  BPOSR~\cite{bposr}, Cai et al.~\cite{cai2024spherical}
            }, color=mycolor_3, fill=mycolor_3, text=black, inner ysep=0.3pt, text width=36.5em
            ]
            [{
                \textbf{\ \,Generative Model-Driven:} OmniSSR~\cite{omnissr}, DiffOSR~\cite{diffosr}, RealOSR~\cite{realosr}
            }, color=mycolor_3, fill=mycolor_3, text=black, inner ysep=0.3pt, text width=36.5em
            ]
            ]
        [Reflection Removal, color=mycolor_3, fill=mycolor_3, text=black, inner ysep=0.3pt, text width=8em, text centered
            [{
                \ \,Hong et al.~\cite{hongrr}, PAR$2$Net~\cite{par2net}, ZS360~\cite{han2022zero}, 
                Park et al.~\cite{park2024fully}
            }, color=mycolor_3, fill=mycolor_3, text=black, inner ysep=0.3pt, text width=36.5em
            ]
        ]
        [Restoration, color=mycolor_3, fill=mycolor_3, text=black, inner ysep=0.3pt, text width=8em, text centered
            [{
                \textbf{\ \,Denoising:} Bigot et al.~\cite{bigot2007spherical}, Iazzi et al.~\cite{ iazzi2014adapted}, Phan et al.~\cite{phan2020space}, SphereDRUNet~\cite{fermanian2023spheredrunet}
            }, color=mycolor_3, fill=mycolor_3, text=black, inner ysep=0.3pt, text width=36.5em
            ]
            [{
                \textbf{\ \,Deblurring:} Li et al.~\cite{li2014omnigradient}, Peng et al.~\cite{peng2012coded},
                Liu et al.~\cite{liu2014coded}, Alibouch et al.~\cite{alibouch2021catadioptric}
            }, color=mycolor_3, fill=mycolor_3, text=black, inner ysep=0.3pt, text width=36.5em
            ]
            [{
                \textbf{\ \,Dehazing:} Zhao et al.~\cite{zhao2023stripe}
            }, color=mycolor_3, fill=mycolor_3, text=black, inner ysep=0.3pt, text width=36.5em
            ]
        ]
        [Quality Assessment, color=mycolor_3, fill=mycolor_3, text=black, inner ysep=0.3pt, text width=8em, text centered
            [{
                \textbf{\ \,Behavior-Aware:} MC360IQA~\cite{mc360iqa}, VGCN~\cite{vgcn}, TVFormer~\cite{yang2022tvformer}, JointNet~\cite{jointnet}, 
                DeepVR-IQA~\cite{deepvriqa}, \\ \ \,Assessor360~\cite{assessor360}, ST360IQ~\cite{st360iq}, VSBNet~\cite{VSBNET}, SCPOIQA~\cite{scpiqa}, CIQNet~\cite{ciqnet}, NR-OVQA~\cite{nrovqa}
            }, color=mycolor_3, fill=mycolor_3, text=black, inner ysep=0.3pt, text width=36.5em
            ]
            [{
                \textbf{\ \,Distortion- and Geometry-Aware:}
                WS-PSNR~\cite{wspsnr},  WS-SSIM~\cite{wsssim}, S-PSNR~\cite{spsnr}, OV-PSNR~\cite{ovpsnr}, 
                CPBQA~\cite{CPBQA}, \\
                \ \,VSBNet~\cite{VSBNET}, 
                V-CNN~\cite{vcnn},
                MP-BOIQA~\cite{MPBOIQA}, 
                360-VQA~\cite{360vqa},
                GAT~\cite{yang2025hierarchical},
                MFILGN~\cite{mfign}, Yang.~\cite{yang2025quality},
                ASAL~\cite{asal}
            }, color=mycolor_3, fill=mycolor_3, text=black, inner ysep=0.3pt, text width=36.5em
            ]
        ]
        [Segmentation, color=mycolor_4, fill=mycolor_4, text=black, inner ysep=0.3pt, text width=8em, text centered
            [{\textbf{\ \,Knowledge Alignment:} DensePASS~\cite{densepass}, DPPASS~\cite{dppass}, Trans4PASS~\cite{trans4pass}, Trans4PASS+~\cite{trans4pass+}, 360SFUDA~\cite{360sfuda}, 360SFUDA++~\cite{360sfuda++}, 
            }, color=mycolor_4, fill=mycolor_4, text=black, inner ysep=0.3pt, inner ysep=0.3pt, text width=36.5em
            ]
            [{\textbf{\ \,Pseudo-Label Generation:} Yang et al.~\cite{yang2020omnisupervised}, GoodSAM~\cite{goodsam}, GoodSAM++~\cite{goodsam++}, OmniSAM~\cite{omnisam}
            }, color=mycolor_4, fill=mycolor_4, text=black, inner ysep=0.3pt, text width=36.5em
            ]
            [{\textbf{\ \,Others:} PASS~\cite{pass}, DS-PASS~\cite{dspass},
            DDCNet~\cite{ddcnet}, Zheng et al.~\cite{zheng2023complementary}, DeepPanoContext~\cite{zhang2021deeppanocontext}, 
            SGAT4PASS~\cite{li2023sgat4pass}, Liu et al. \cite{liu2025360}
            HexRUNet~\cite{zhang2019orientation} \\ \ \,PRF~\cite{prf}, PanoVOS~\cite{Panovos}, Pano-SfMLearner~\cite{Pano-SfMLearner}, OASS~\cite{oass}, OOOPS~\cite{ooops} 
            }, color=mycolor_4, fill=mycolor_4, text=black, inner ysep=0.3pt, text width=36.5em
            ]   
        ]
        [Mapping, color=mycolor_4, fill=mycolor_4, text=black, inner ysep=0.3pt, text width=8em, text centered
            [{
                \ \,360Mapper~\cite{360bev}, OneBEV~\cite{onebev}, HumanoidPano~\cite{zhang2025humanoidpano}
            }, color=mycolor_4, fill=mycolor_4, text=black, inner ysep=0.3pt, inner ysep=0.3pt, text width=36.5em
            ]
        ]
        [Detection, color=mycolor_4, fill=mycolor_4, text=black, inner ysep=0.3pt, text width=8em, text centered
            [{\textbf{\ \,Distortion-aware:} SphereNet~\cite{spherenet}, SPHCONV~\cite{sphconv}, Curved-Space Faster R-CNN~\cite{curvedspacercnn}, MS-RPN~\cite{msrpn}, PDAT~\cite{pdat}, PanoGlassNet~\cite{chang2024panoglassnet}
            }, color=mycolor_4, fill=mycolor_4, text=black, inner ysep=0.3pt, text width=36.5em
            ]
            [{\textbf{\ \,Projection-Driven:} Multi-Projection YOLO~\cite{yang2018object}, Rep R-CNN~\cite{reprcnn}
            }, color=mycolor_4, fill=mycolor_4, text=black, inner ysep=0.3pt, inner ysep=0.3pt, text width=36.5em
            ]
            [{\textbf{\ \,Redefining spherical bounding:} Unbiased IoU~\cite{unbiasediou}, FoV-IoU~\cite{foviou}, Sph2Pob~\cite{sph2pob}
            }, color=mycolor_4, fill=mycolor_4, text=black, inner ysep=0.3pt, text width=36.5em
            ]
        ]
        [Tracking, color=mycolor_4, fill=mycolor_4, text=black, inner ysep=0.3pt, text width=8em, text centered
            [{
                \ \,Jiang et al.~\cite{jiangtracking}, MMPAT~\cite{mmpat}, CC3DT~\cite{cc3dt}, 360VOT~\cite{360vot}, 360VOTS~\cite{360vots}, Luo et al. \cite{luo2025omnidirectional}
                }, color=mycolor_4, fill=mycolor_4, text=black, inner ysep=0.3pt, inner ysep=0.3pt, text width=36.5em
            ]
        ]
        [Pose Estimation, color=mycolor_4, fill=mycolor_4, text=black, inner ysep=0.3pt, text width=8em, text centered
            [{
                \ \,CoVisPose~\cite{covispose}, Graph-CoVis~\cite{Graph-CoVis}, PanoPose~\cite{panopose}
            }, color=mycolor_4, fill=mycolor_4, text=black, inner ysep=0.3pt, inner ysep=0.3pt, text width=36.5em
            ]
        ]
        [Saliency Prediction, color=mycolor_4, fill=mycolor_4, text=black, inner ysep=0.3pt, text width=8em, text centered
            [{
                \textbf{\ \,Projection-Driven:} SalNet~\cite{monroy2018salnet360}, Suzuki et al~\cite{ suzuki2018saliency}, Dai et al~\cite{dai2020dilated}, Djemai et al~\cite{djemai2020extending}, SalGAN360~\cite{chao2018salgan360}, SalbiNet~\cite{chen2020salbinet360}, Dedhia et al.~\cite{dedhia2019saliency}
            }, color=mycolor_4, fill=mycolor_4, text=black, inner ysep=0.3pt, text width=36.5em
            ]
            [{
                \textbf{\ \,Spherical GNN-based:} SalGCN~
                    \cite{salgcn}, SalReGCN360~\cite{salregcn360},
                    SalGFCN~\cite{yang2021salgfcn}, GBCNN~\cite{zhu2021viewing},
                    360Spred~\cite{yang2024360spred}
            }, color=mycolor_4, fill=mycolor_4, text=black, inner ysep=0.3pt, text width=36.5em
            ]
            [{
                \textbf{\ \,Behavior Modeling:} Abreu et al.~\cite{de2017look}, Zhu et al.~\cite{zhu2018prediction}, Bur et al.~\cite{bur2006robot}, 
                PanoSalNet~\cite{panosalnet}, \\ \ \,Qiao et al.~\cite{viewport},
                Cheng et al.~\cite{cubepadding}
                DHP~\cite{dhp},
                ATSal~\cite{atsal},
                Guo et al.~\cite{guo2024instance}, 
                Cokelek et al.~\cite{cokelek2025spherical}
            }, color=mycolor_4, fill=mycolor_4, text=black, inner ysep=0.3pt, text width=36.5em
            ]
            [{
                \textbf{\ \,Others:} Zhu et al.~\cite{zhu2021lightweight}, RANSP~\cite{ransp, zhu2021saliency}, Abreu et al.~\cite{de2017look}, Zhu et al. \cite{zhu2018prediction},  Bogdanova et al.~\cite{bogdanova2008visual}
                }, color=mycolor_4, fill=mycolor_4, text=black, inner ysep=0.3pt, text width=36.5em
            ]
        ]
        [Layout Detection, color=mycolor_4, fill=mycolor_4, text=black, inner ysep=0.3pt, text width=8em, text centered
            [{
                \textbf{\ \,Projection-Driven}: DuLa-Net~\cite{dulanet}, PSMNet~\cite{wang2022psmnet}, uLayout~\cite{ulayout}
                }, color=mycolor_4, fill=mycolor_4, text=black, inner ysep=0.3pt, text width=36.5em
            ]
            [{\textbf{\ \,Geometric-Based:} HorizonNet\cite{horizonnet}, DOPNet~\cite{dopnet}, MV-DOPNet~\cite{shen2024360}, C2P-Net~\cite{zhang2025c2p}, LED$^2$-Net~\cite{wang2021led}, Seg2Reg~\cite{sun2024seg2reg}, LGT-Net~\cite{lgtnet}, HoHoNet~\cite{hohonet}
            }, color=mycolor_4, fill=mycolor_4, text=black, inner ysep=0.3pt, text width=36.5em
            ]
            [{\textbf{\ \,Structural-Aware:} Fernandez et al.~\cite{fernandez2018layouts}, PanoContext~\cite{zhang2014panocontext}, GPRNet~\cite{gprnet}, Bi-Layout~\cite{tsai2024no}, Jia et al.~\cite{jia20223d}, SSLayout360~\cite{tran2021sslayout360}, SemiLayout360~\cite{zhang2025semi}
            }, color=mycolor_4, fill=mycolor_4, text=black, inner ysep=0.3pt, text width=36.5em
            ]
        ]
        [Optical Flow Estimation, color=mycolor_4, fill=mycolor_4, text=black, inner ysep=0.3pt, text width=8em, text centered
            [{\textbf{\ \,Distortion-Aware:} Cubes3D~\cite{apitzsch2018cubes3d}, SimpleNet~\cite{xie2019effective}, OmniFlowNet~\cite{artizzu2021omniflownet}, LiteFlowNet360~\cite{bhandari2021revisiting}, PanoFlow~\cite{panoflow},
            PriOr-Flow~\cite{liu2025prior}
            }, color=mycolor_4, fill=mycolor_4, text=black, inner ysep=0.3pt, inner ysep=0.3pt, text width=36.5em
            ]
            [{\textbf{\ \,Projection-Driven:} Yuan et al.~\cite{yuan2021360},
            Li et al.~\cite{li2022deep}
            }, color=mycolor_4, fill=mycolor_4, text=black, inner ysep=0.3pt, text width=36.5em
            ]
        ]
        [Keypoint Matching, color=mycolor_4, fill=mycolor_4, text=black, inner ysep=0.3pt, text width=8em, text centered
            [{
                \ \,SPHORB~\cite{zhao2015sphorb}, Chuang et al.~\cite{chuang2018rectified}, PanoPoint~\cite{panopoint}, SphereGlue~\cite{sphereglue}, EDM~\cite{edm}
            }, color=mycolor_4, fill=mycolor_4, text=black, inner ysep=0.3pt, inner ysep=0.3pt, text width=36.5em
            ]
        ]
        [Decomposition, color=mycolor_4, fill=mycolor_4, text=black, inner ysep=0.3pt, text width=8em, text centered
            [{
                \ \,Li et al.\cite{li2021lighting}, PhyIR\cite{li2022phyir}, Xu et al.~\cite{xu2024intrinsic}, 
            }, color=mycolor_4, fill=mycolor_4, text=black, inner ysep=0.3pt, inner ysep=0.3pt, text width=36.5em
            ]
        ]
        [Lighting Estimation, color=mycolor_4, fill=mycolor_4, text=black, inner ysep=0.3pt, text width=8em, text centered
            [{\textbf{\ \,Physics-Based:} Weber et al.~\cite{weber2018learning}, Gkitsas et al.~\cite{gkitsas2020deep}, Hold et al.~\cite{hold2017deep, hold2019deep}, Zhang et al.~\cite{zhang2019all}, 
            Song et al.~\cite{song2019neural}, Garon et al.~\cite{garon2019fast}
            }, color=mycolor_4, fill=mycolor_4, text=black, inner ysep=0.3pt, inner ysep=0.3pt, text width=36.5em
            ]
            [{\textbf{\ \,Panorama Generation:} EnvMapNet~\cite{somanath2021hdr}, StyleLight~\cite{wang2022stylelight}, IllumiDiff~\cite{shen2025illumidiff}, 
            Hilliard et al.~\cite{hilliard2025hdr},
            EMLight~\cite{zhan2021emlight}, GMLight~\cite{zhan2022gmlight} \\
            \ \,Weber et al.~\cite{weber2022editable}, EverLight~\cite{dastjerdi2023everlight},
            SOLID-Net~\cite{zhu2021spatially}, SALENet~\cite{zhao2024salenet},
            CleAR~\cite{zhao2024clear}
            }, color=mycolor_4, fill=mycolor_4, text=black, inner ysep=0.3pt, text width=36.5em
            ]
        ]
        [Depth Estimation, color=mycolor_4, fill=mycolor_4, text=black, inner ysep=0.3pt, text width=8em, text centered
            [{\textbf{\ \,Distortion-Aware:} OmniDepth~\cite{zioulis2018omnidepth}, Tateno et al.~\cite{tateno2018distortion}, ACDNet~\cite{zhuang2022acdnet}, PanoFormer~\cite{shen2022panoformer}, EGFormer~\cite{yun2023egformer}, OmniDiffusion~\cite{mohadikar2025omnidiffusion}, HUSH~\cite{lee2025hush}
            }, color=mycolor_4, fill=mycolor_4, text=black, inner ysep=0.3pt, inner ysep=0.3pt, text width=36.5em
            ]
            [{\textbf{\ \,Projection-Driven:} BiFuse~\cite{wang2020bifuse}, UniFuse~\cite{jiang2021unifuse},
            Peng et al.~\cite{peng2023high}, GLPanoDepth~\cite{bai2024glpanodepth}, HRDFuse~\cite{ai2023hrdfuse}, 
            Ai et al.~\cite{ai2024elite360d, ai2024elite360m} \\ \ \,OmniFusion~\cite{li2022omnifusion}, 360MonoDepth~\cite{rey2022360monodepth}, SphereUFormer~\cite{benny2025sphereuformer}, S$^{2}$Net\cite{li2023mathcal}, MS360~\cite{ mohadikar2024ms360}, SGFormer~\cite{zhang2025sgformer}, PGFuse~\cite{shen2024revisiting}, OmniStereo~\cite{deng2025omnistereo}
            }, color=mycolor_4, fill=mycolor_4, text=black, inner ysep=0.3pt, text width=36.5em
            ]
            [{\textbf{\ \,Others:} Slicenet~\cite{pintore2021slicenet} 
            HoHoNet~\cite{sun2021hohonet}, PanelNet~\cite{yu2023panelnet}, Huang et al.
            ~\cite{huang2024multi}, Pano Popups~\cite{eder2019pano}, Feng et al.~\cite{feng2020deep}, Zioulis et al.~\cite{zioulis2019spherical}, Wang et al.~\cite{wang2023distortion} \\ \ \,Yun et al. \cite{yun2022improving}, Spdet~\cite{zhuang2023spdet}, Garg et al.~\cite{garg2016unsupervised}, Depth Anywhere~\cite{wang2024depth}, PanDA~\cite{cao2025panda}, Unik3D~\cite{piccinelli2025unik3d}, Depth Any Camera~\cite{guo2025depth}
            }, color=mycolor_4, fill=mycolor_4, text=black, inner ysep=0.3pt, text width=36.5em
            ]
        ]
        [Fusion Modeling with Audio, color=mycolor_6, fill=mycolor_6, text=black, inner ysep=0.3pt, text width=8em, text centered
            [{\ \,Audible Panorama~\cite{huang2019audible}, Li et al.\cite{li2018scene}, Morgado et al.\cite{morgado2018self}, AVS-ODV~\cite{zhu2023audio}, PAV-SOD~\cite{zhang2023pav}, PAV-SOR~\cite{guo2024instance}\\ \ \,Pano-AVQA~\cite{yun2021pano}, Masuyama et al.~\cite{masuyama2020self}, Vasudevan et al.~\cite{vasudevan2020semantic}, OAVQA~\cite{zhu2023perceptual},
            Fela et al.~\cite{fela2022perceptual}, 
            Sonic4D~\cite{xie2025sonic4d}
            }, color=mycolor_6, fill=mycolor_6, text=black, inner ysep=2pt, text width=36.5em   
            ]
        ]
        [Fusion Perception with LiDAR, color=mycolor_6, fill=mycolor_6, text=black, inner ysep=0.3pt, text width=8em, text centered  
            [{\ \,SPOMP~\cite{miller2024air}, HumanoidPano~\cite{zhang2025humanoidpano}, Ma et al. \cite{ma2020method}, OmniColor \cite{liu2024omnicolor}, Zhao et al.~\cite{zhao2022attention}, Bernreiter et al. \cite{bernreiter2021spherical}, PanoramicVO \cite{yuan2024panoramic}, MMPAT~\cite{mmpat}
            }, color=mycolor_6, fill=mycolor_6, text=black, inner ysep=2pt, text width=36.5em   
            ]
        ]
        [Fusion with Text, color=mycolor_6, fill=mycolor_6, text=black, inner ysep=0.3pt, text width=8em, text centered
                [{\ \,VQA-360~\cite{chou2020visual}, VIEW-QA~\cite{song2024video}, OmniVQA~\cite{zhang2025towards}, OSR-Bench~\cite{dongfang2025multimodal}, Dense360~\cite{zhou2025dense360}
                }, color=mycolor_6, fill=mycolor_6, text=black, inner ysep=2pt, text width=36.5em   
                ]
        ]        
        [Text-guided Generation,  color=mycolor_5, fill=mycolor_5, text=black, inner ysep=0.3pt, text width=8em, text centered
            [{
                \textbf{\ \,Distortion-Aware:} Text2Light~\cite{chen2022text2light}, 
                SMGD~\cite{sun2025spherical},
                SphereDiffusion~\cite{wu2024spherediffusion}, 
                PanoWAN~\cite{xia2025panowan},
                DynamicScaler~\cite{liu2025dynamicscaler},
                PanoDiT~\cite{zhang2025panodit}
            }, color=mycolor_5, fill=mycolor_5, text=black, inner ysep=0.3pt, text width=36.5em
            ]
            [{
                \textbf{\ \,Projection-Driven:} TanDiT~\cite{ccapuk2025tandit},
                Zhang et al.\cite{zhang2024taming},
                TiP4GEN~\cite{xing2025tip4gen},
                DreamCube~\cite{huang2025dreamcube},
                SphereDiff\cite{park2025spherediff},
                VideoPanda~\cite{xie2025videopanda}, ViewPoint~\cite{fang2025panoramic}
            }, color=mycolor_5, fill=mycolor_5, text=black, inner ysep=0.3pt, text width=36.5em
            ]
            [{
                \textbf{\ \,Continuity Modeling:} Diffusion360~\cite{feng2023diffusion360}, StitchDiffusion~\cite{wang2024customizing}, PAR~\cite{wang2025conditional}, PanoFree~\cite{liu2024panofree}
            }, color=mycolor_5, fill=mycolor_5, text=black, inner ysep=0.3pt, text width=36.5em
            ]
            [{
                \textbf{\ \,Others:} 
                VidPanos~\cite{ma2024vidpanos},
                OmniDrag~\cite{li2024omnidrag}
            }, color=mycolor_5, fill=mycolor_5, text=black, inner ysep=0.3pt, text width=36.5em
            ]
        ]
        [Image Completion, color=mycolor_5, fill=mycolor_5, text=black, inner ysep=0.3pt, text width=8em, text centered
            [{\textbf{\ \,Distortion-Aware:} Dream360~\cite{ai2024dream360}, PanoDecouple~\cite{zheng2025panorama}, 
            2S-ODIS~\cite{nakata20242s}
            }, color=mycolor_5, fill=mycolor_5, text=black, inner ysep=0.3pt, inner ysep=0.3pt, text width=36.5em
            ]
            [{\textbf{\ \,Continuity Modeling:} 
            Cylin-Painting~\cite{liao2023cylin},
            PanoDiff~\cite{wang2023360},
            PanoDiffusion~\cite{wu2023panodiffusion},
            Akimoto et al.'s ~\cite{akimoto2022diverse}
            }, color=mycolor_5, fill=mycolor_5, text=black, inner ysep=0.3pt, text width=36.5em
            ]
            [{\textbf{\ \,Others:} 
            ImmerseGAN~\cite{dastjerdi2022guided}, AOG-Net~\cite{lu2024autoregressive}, 
            BIPS~\cite{oh2022bips}
            }, color=mycolor_5, fill=mycolor_5, text=black, inner ysep=0.3pt, text width=36.5em
            ]
        ]
        [Novel View Synthesis, color=mycolor_5, fill=mycolor_5, text=black, inner ysep=0.3pt, text width=8em, text centered
            [{\textbf{\ \,NeRF-based:} EgoNeRF~\cite{choi2023balanced}, PanoGRF~\cite{chen2023panogrf}, 
            Omni-NeRF~\cite{gu2022omni}, 
            OmniLocalRF~\cite{choi2024omnilocalrf}, 
            360Roam~\cite{huang2022360roam}, 360FusionNeRF~\cite{kulkarni2023360fusionnerf}, PERF~\cite{wang2024perf}, PanoHDR-NeRF~\cite{gera2022casual}
            }, color=mycolor_5, fill=mycolor_5, text=black, inner ysep=0.3pt, inner ysep=0.3pt, text width=36.5em
            ]
            [{\textbf{\ \,3DGS-based:} 360-GS~\cite{bai2024360}, ODGS~\cite{lee2024odgs}, PanSplat~\cite{zhang2025pansplat}, OmniGS~\cite{li2024omnigs}, Splatter-360~\cite{chen2025splatter}, OmniSplat~\cite{lee2025omnisplat} \\ 
            \ \,TPGS~\cite{shen2025you}, PanoSplatt3R~\cite{ren2025panosplatt3r}, ErpGS~\cite{ito2025erpgs}, Seam360GS~\cite{shin2025seam360gs},  OB3D~\cite{ito2025ob3d}
            }, color=mycolor_5, fill=mycolor_5, text=black, inner ysep=0.3pt, text width=36.5em
            ]
            [{\textbf{\ \,Lightweight Panoramic View Synthesis:} 
                OmniSyn~\cite{li2022omnisyn}, SOMSI~\cite{habtegebrial2022somsi}, Casual 6-DoF~\cite{chen2022casual}
            }, color=mycolor_5, fill=mycolor_5, text=black, inner ysep=0.3pt, text width=36.5em
            ]
        ]
        [Generation-Driven Applications, color=mycolor_5, fill=mycolor_5, text=black, inner ysep=0.3pt, text width=8em, text centered
            [{
                \textbf{\ \,Constrained Scene Generation:} SceneDreamer360~\cite{li2024scenedreamer360}, DreamScene360~\cite{zhou2024dreamscene360}, LayerPano3D~\cite{yang2024layerpano3d}, HoloDreamer~\cite{zhou2024holodreamer}, HoloTime~\cite{zhou2025holotime}, 4K4DGen~\cite{li20244k4dgen}
            }, color=mycolor_5, fill=mycolor_5, text=black, inner ysep=0.3pt, text width=36.5em
            ]
            [{
                \textbf{\ \,Unconstrained World Generation:} 
                            HunyuanWorld 1.0~\cite{team2025hunyuanworld},
                            Matrix-3D~\cite{yang2025matrix}
            }, color=mycolor_5, fill=mycolor_5, text=black, inner ysep=0.3pt, text width=36.5em
            ]
            [{
                \textbf{\ \,Vision–Language Navigation (VLN):} PANOGEN~\cite{li2023panogen}, PANOGEN++\cite{wang2025panogen++}, VLN-RAM~\cite{wei2025unseen}, Any2Omni~\cite{yang2025omni} 
                }, color=mycolor_5, fill=mycolor_5, text=black, inner ysep=0.3pt, text width=36.5em
            ]
        ]
    ]
\end{forest}
}
\caption{Summary of Panoramic Vision Tasks and Representative Methods.}
\vspace{-15pt}
\label{tab:tasks}
\end{figure*}
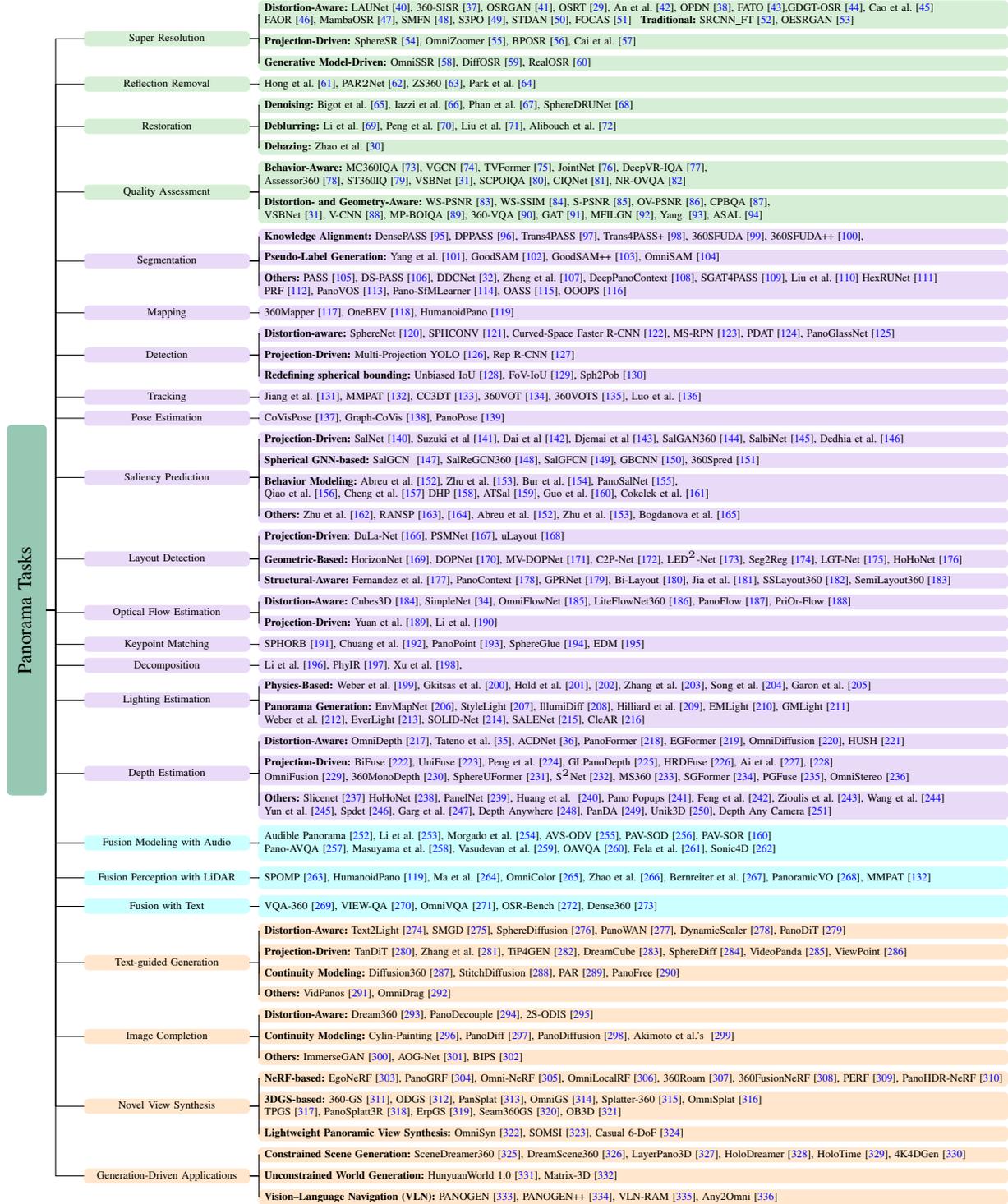

\section{Panorama Tasks}
\label{sec:pt}

Recent advances in panoramic vision have catalyzed a wide range of tasks across perception, understanding, and generation. As summarized in Table~\ref{tab:tasks}, over 20 representative tasks across four categories have been explored, ranging from low-level image enhancement to high-level scene understanding, multimodal fusion, and immersive content generation. This section gives a comprehensive review of these tasks, conducting a horizontal, cross-task analysis of the methods proposed to address the unique gaps introduced by panoramic scenarios.

\subsection{Visual Quality Enhancement and Assessment}
The real-world images typically suffer quality degradation due to various disturbances encountered during compression, transmission, and acquisition processes~\cite{lin2023unsupervised,lin2024dual,lin2025re}.
The visual quality of images significantly affects human perceptual experiences as well as the performance of subsequent downstream tasks, such as image segmentation, object detection, and 3D reconstruction. 
To address these issues, in addition to the imaging system-based solutions mentioned in Section \ref{sec:is}, image quality enhancement serves as a post-processing approach after image acquisition. It aims to restore high-quality, high-fidelity images from degraded inputs by emphasizing fine details and improving the overall visual clarity. 
Concurrently, image quality assessment provides quantitative evaluations of both degraded and enhanced images through objective or subjective metrics, guiding the development and optimization of enhancement methodologies.

In particular, perspective images have witnessed remarkable progress in quality enhancement and assessment, benefiting from a range of learning-based techniques.
However, panoramic images exhibit distinctive characteristics compared to perspective images, such as severe distortions near the poles and uneven pixel distribution. These features pose substantial barriers to the direct application of existing perspective-based techniques.
Therefore, this section systematically reviews recent advancements in panoramic image quality enhancement and assessment, identifying critical technical trends and highlighting open research challenges within this emerging field.
Since Section \ref{subsec:method comparison} provides a detailed analysis of the advantages, limitations, and applicability of the two common strategies, this section presents only a brief discussion for the strategies. For task-specific future directions, please refer to the supplementary material.

\subsubsection{Super Resolution} 

Super-resolution aims to reconstruct a high-resolution image or video from one or more low-resolution inputs by restoring fine details and enhancing visual quality. Some early methods~\cite{SRCNNFT, oesrgan} extend the existing perspective-based models~\cite{SRCNN, esrgan} with panorama data.
However, their performance is significantly constrained by model designs that fail to account for the unique characteristics distinguishing panoramic from perspective ones.
To address these issues, advanced methods can be categorized into three groups:

\noindent \textbf{Distortion-Aware Methods} are designed to address the non-uniform pixel distribution~\cite{launet} in ODI-SR by introducing distortion-aware priors or adaptive weighting schemes that emphasize perceptually important equatorial regions. 
The proposed methods with hierarchical modeling (LAUNet~\cite{launet}), distortion-aware priors (360-SISR~\cite{360sisr}), weighted loss designs (OSRGAN~\cite{osrgan}), distortion-aware convolutions and adaptive losses (OSRT~\cite{osrt}, An et al.\cite{an2023perception}, Sun et al.\cite{sun2023opdn}), pixel-wise weighting and distortion-guided attention (FATO~\cite{an2024fato}, GDGT-OSR~\cite{gdgtosr}), and more advanced transformer-based schemes leveraging geometric, semantic, and frequency cues (Cao et al.\cite{cao2025geometric}, Shen et al.\cite{shen2025fast}, Wen et al.~\cite{wen2025mambaosr}).
For video SR, works such as SMFN~\cite{smfn}, S3PO~\cite{s3po}, and STDAN~\cite{an2024spatio} adopt latitude-aware losses to emphasize equatorial regions and temporal-aware modules for spatiotemporal coherence. Central vision-based SR has also been explored in FOCAS~\cite{wang2023central}, leveraging foveated rendering to highlight human central vision.

\noindent \textbf{Projection-Driven Methods} leverage alternative projections to reduce ERP distortions, enabling super-resolution models to better pursue local geometric precision and structural fidelity. Some works include continuous spherical modeling with spherical CNNs on icosahedral grids (SphereSR~\cite{spheresr}), Möbius projection with spatially adaptive resampling for localized high-precision upsampling (OmniZoomer~\cite{omnizoomer}), dual-branch geometric alignment for structural consistency (BPOSR~\cite{bposr}), and pseudo-cylindrical representations for adaptive latitude sampling compatible with standard 2D SR networks (Cai et al.~\cite{cai2024spherical}).

\noindent \textbf{Generative Model-Driven Approaches} leverage the strong priors of foundation models, particularly diffusion models, to handle unknown and complex degradations and improve generalization in ODISR. 
Some works include ERP-to-TP projection interaction with gradient decomposition correction for detail recovery (OmniSSR~\cite{omnissr}), stepwise sampling to approximate high-resolution distributions and reduce texture blurring (DiffOSR~\cite{diffosr}), and efficient realistic SR with single-step sampling and unfolding-guided injector for complex degradations (RealOSR~\cite{realosr}).

Overall, distortion-aware methods emphasize global consistency and perceptual quality, while projection-driven methods prioritize local geometric fidelity. In contrast, generative model-driven approaches remain at an early stage; they still suffer from high computational cost and slow inference.

\begin{figure*}[t]
\centering
\includegraphics[width=1\linewidth]{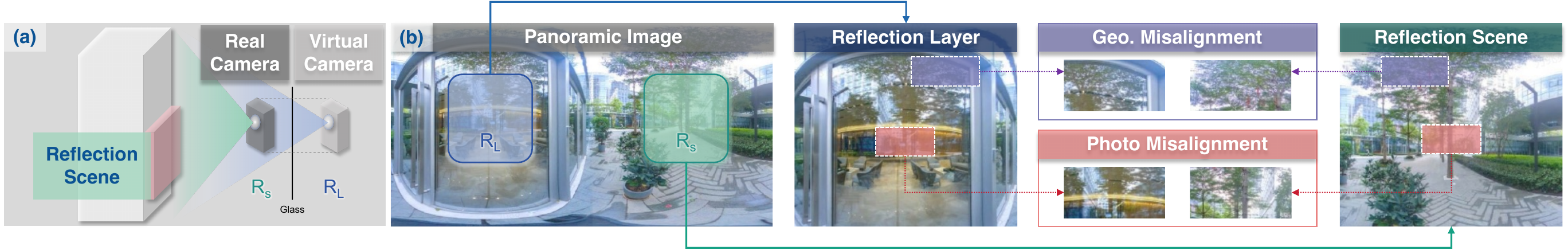}
\vspace{-15pt}
\caption{Reflection removal challenges in panoramic images. The reflection scene ($R_s$, green) and reflection layer ($R_l$, blue) present geometric misalignment and photo misalignment, which make transmission-reflection separation more difficult.}
\vspace{-15pt}
\label{fig:reflection}
\end{figure*}

\subsubsection{Reflection Removal.}

Reflection removal aims to recover the true transmission scene from glass-mixed inputs, but traditional perspective-based methods often fail when applied to panoramic images where reflections can be sharp and complex, as shown in Fig.~\ref{fig:reflection}. 
Most perspective-based methods assume the whole image is glass-mixed with weak and blurred reflections, which often fails in 360 panoramas, making transmission–reflection separation particularly challenging.

Recent panorama-based approaches leverage the unique characteristic that both the reflective scene and the mixture image within the same frame: introducing external priors from visible reflection sources (Hong et al.~\cite{hongrr}), unifying correspondence modeling, transmission recovery, and reflection refinement in end-to-end designs (Par$^{2}$Net~\cite{par2net}), extending to zero-shot reflection removal via iterative geometric matching and disentanglement (ZS360~\cite{han2022zero}), and developing fully automated frameworks with long-range dependency modeling and multi-scale alignment for robust performance (Park et al.~\cite{park2024fully}).

\subsubsection{Omnidirectional Image Restoration}
Omnidirectional image restoration aims to restore high-quality 360\textdegree{} images from their degraded counterparts, which can include noise, blur, and weather distortions. 
Here, we review methods for main image restoration tasks, including denoising, deblurring, and dehazing.
In the following sections, we will highlight how each approach leverages the spatial geometry of omnidirectional data to achieve robust performance.

\noindent \textit{\underline{(1) Image Denoising}}: Early methods adapt classical filters to spherical geometry, including Wiener filtering, Tikhonov regularization, and Stein block thresholding~\cite{bigot2007spherical, iazzi2014adapted}; panorama-adapted designs introduce space-variant total variation to model ERP-specific distortions~\cite{phan2020space}; and recent learning-based approaches like SphereDRUNet leverage uniform HEALPix sampling to perform data-driven restoration directly on the sphere~\cite{fermanian2023spheredrunet}.

\noindent \textit{\underline{(2) Image deblurring}}: Only some traditional methods are proposed for omnidirectional images, adapting classical strategies to address geometric and optical challenges. Some approaches include projection-based priors such as Omnigradient, which incorporates cylindrical gradient regularization into deconvolution~\cite{li2014omnigradient}; hardware-based solutions like coded apertures to capture all-focus information and mitigate defocus blur~\cite{peng2012coded, liu2014coded}; and spherical-domain filtering methods such as harmonic-based Wiener filtering on the 2-sphere~\cite{alibouch2021catadioptric}.

\noindent \textit{\underline{(3) Image dehazing}}:
To address panoramic dehazing, Zhao et al.~\cite{zhao2023stripe} propose a distortion-aware convolution to handle ERP-induced distortion. Their end-to-end framework jointly performs dehazing and depth estimation, establishing a strong baseline for adverse-weather restoration in omnidirectional scenes.

\subsubsection{Visual Quality Assessment.} 

Visual Quality Assessment aims to quantitatively evaluate the perceptual quality of panoramic images and videos. 
Based on whether reference data are used as constraints, these methods can be categorized into full reference (FR) and no-reference (NR) approaches. 
Unlike conventional IQA and VQA, which often assume uniform visibility and regress a single global score, panoramic quality assessment faces unique challenges arising from ERP distortions near the poles and from user-dependent viewports that expose only localized regions at a time. 
To address these issues, two representative strategies have recently emerged.

\noindent \textbf{Behavior-Aware Methods} aim to reflect human subjective perception by simulating localized viewing on head-mounted displays and integrating behavioral cues such as eye movement, head movement, and saliency. Some works include shared CNNs and graph reasoning for inter-viewport relationships (MC360IQA~\cite{mc360iqa}, VGCN~\cite{vgcn}); sequence models for temporal scanning and memory effects (TVFormer~\cite{yang2022tvformer}, JointNet~\cite{jointnet}); behavioral priors such as viewing coordinates, adversarial learning, and trajectory/saliency-guided weighting (DeepVR-IQA~\cite{deepvriqa}, Assessor360~\cite{assessor360}, ST360IQ~\cite{st360iq}). Recently, fidelity-enhanced designs have been proposed to align distorted content with pseudo-references or to fuse predicted saliency into score aggregation (VSBNet~\cite{VSBNET}, SCPIQA~\cite{scpiqa}). For VQA, perceptual- and causal-aware models for robust quality estimation (CIQNet~\cite{ciqnet}, NR-OVQA~\cite{nrovqa}). While effective under uniformly distributed distortions, these methods still struggle to handle the spatially complex and uneven distortion patterns.

\noindent \textbf{Distortion- and Geometry-Aware Methods} explicitly address spatial non-uniformity and geometric distortions through latitude-aware designings (WS-PSNR~\cite{wspsnr}, WS-SSIM~\cite{wsssim}, S-PSNR~\cite{spsnr}, OV-PSNR~\cite{ovpsnr}); saliency- and viewport-weighted assessment with equidistant convolutions for perceptual fidelity (CPBQA~\cite{CPBQA}, VSBNet~\cite{VSBNET}, V-CNN~\cite{vcnn}); multi-projection and statistical or dynamic modeling for distortion distribution fitting (MP-BOIQA~\cite{MPBOIQA}, 360-VQA~\cite{360vqa}); hierarchical graph attention module (GAT~\cite{yang2025hierarchical}), weakly supervised frequency domain evaluation via wavelet decomposition and NSS statistics (MFILGN~\cite{mfign}); and large-scale benchmarks with BLIP-2-based modeling to jointly predict perceptual quality and degraded regions for AIGC content (Yang et al.~\cite{yang2025quality}), and continual learning approaches tackling cross-dataset generalization and catastrophic forgetting (ASAL~\cite{asal}).

\begin{figure*}[t]
\centering
\includegraphics[width=1\linewidth]{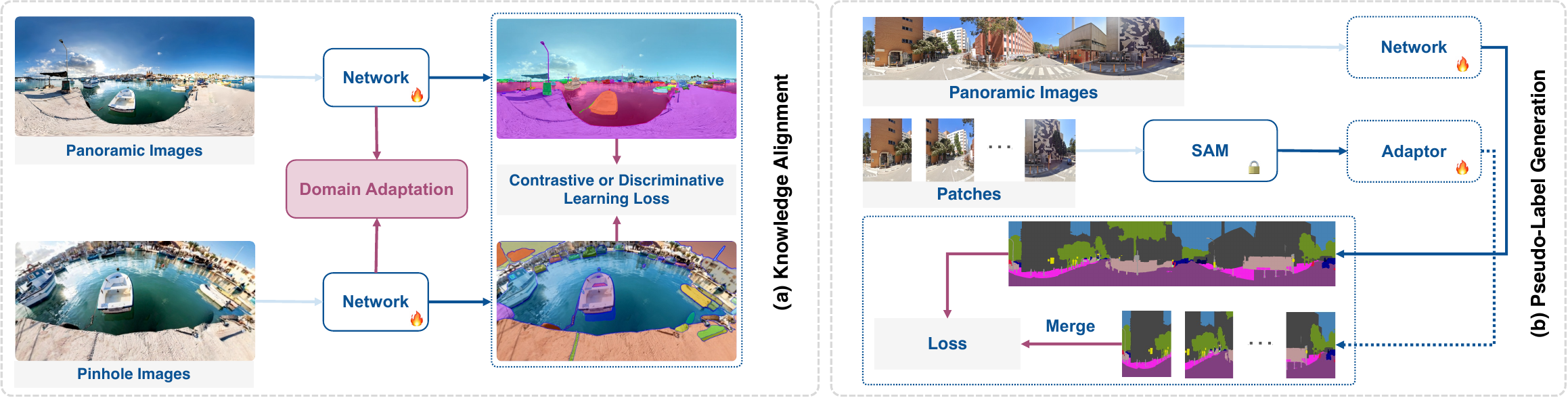}
\caption{UDA strategies for panoramic segmentation. (a) Knowledge Alignment: enforces semantic consistency via adversarial learning and prototypical adaptation. (b) Pseudo-Label Generation: derives supervision from ensemble predictions, SAM-based rectification, and fine-tuning with FoV-aware prototypes and dynamic labeling.}
\vspace{-10pt}
\label{fig:mapping}
\end{figure*}

\subsection{Visual Understanding}

Panoramic visual understanding tasks encompass a wide range of objectives, including high-level semantic interpretation and geometry-centric structural perception. 
To provide a clear task definition and methodological developments, we categorize existing tasks into two groups: (1) \textbf{Semantic-Level} (e.g., object segmentation, detection, saliency prediction, visual tracking), which emphasize region- or pixel-level recognition based on semantic priors; and (2) \textbf{Structure and Motion-Oriented} (e.g., layout, optical flow, pose, lighting, and keypoint estimation), which focus on 3D structure, spatial layout, motion, and illumination. Unlike low-level enhancement, these tasks require higher-level perception, where spatial consistency, geometric invariance, and semantic awareness are critical. However, panoramic distortions and view-dependent semantics pose fundamental challenges that drive the need for specialized representations and learning strategies.

\subsubsection{Object Segmentation}

Object segmentation aims to segment key regions in an image by assigning category labels and, in some cases, instance identities to each pixel. 
While perspective segmentation~\cite{qi2022open} has achieved remarkable progress with large-scale datasets and deep neural networks, panoramic images represented in ERP format introduce new challenges due to polar distortions and violations of planar assumptions. Early methods project ERP images into multiple perspective-like patches and segment them individually before fusion (PASS~\cite{pass}, DS-PASS~\cite{dspass}), which suffers from loss of global context, boundary inconsistencies, and redundant computation. 
To overcome these limitations and reduce reliance on scarce panoramic annotations, recent studies have explored Unsupervised Domain Adaptation (UDA) to transfer knowledge from perspective to panorama. Existing approaches are broadly categorized into two groups, as shown in Fig.~\ref{fig:mapping}.

\noindent \textbf{Knowledge Alignment} focuses on transferring semantic consistency from the source perspective domain to the target panoramic domain, through explicit or implicit alignment mechanisms.
Explicit strategies enforce domain invariance via adversarial learning, focusing on global–local consistency (DensePASS~\cite{densepass}) or aligning ERP and perspective branches for cross-domain generalization (DPPASS~\cite{dppass}). Implicit strategies adopt prototypical adaptation, such as self-supervised Mutual Prototypical Adaptation (Trans4PASS~\cite{trans4pass}) and Segment Anything Model (SAM)~\cite{sam} enhanced prototype correction for supervision (Trans4PASS+\cite{trans4pass+}). To further mitigate projection distortion and style inconsistencies, multi-projection fusion approaches (360SFUDA\cite{360sfuda}, 360SFUDA++~\cite{360sfuda++}) integrate ERP, FFP, and TP views for multi-level alignment of predictions, prototypes, and features, effectively bridging semantic, geometric, and stylistic gaps.

\noindent \textbf{Pseudo-Label Generation} produces supervision signals for unlabeled panoramic images by using pretrained models from other domains. Early approaches generate multiple predictions from transformed panoramic inputs and ensemble them into pseudo-labels (Yang et al.~\cite{yang2020omnisupervised}); SAM-based methods leverage zero-shot masks refined with Teacher Assistant guidance and distortion-aware rectification for improved accuracy at lower cost (GoodSAM~\cite{goodsam}, GoodSAM++~\cite{goodsam++}); and recent advances fine-tune SAM2~\cite{sam2} with LoRA layers, introducing FoV-based prototypical adaptation and dynamic pseudo-labeling to handle distortion, incompleteness, and domain gaps (OmniSAM~\cite{omnisam}). Overall, leveraging the strong generalization ability of large pretrained models provides a promising paradigm for more reliable panoramic segmentation under limited supervision.

\noindent \textbf{Others.} There are also some more specialized designs: distortion convolution~\cite{ddcnet}, bi-directional learning module~\cite{zheng2023complementary}, multi-task learning~\cite{zhang2021deeppanocontext}, spherical deformable embedding~\cite{li2023sgat4pass}, dual-branch designing~\cite{liu2025360}, and unfolded icosahedron mesh~\cite{zhang2019orientation}.
In addition, with the development of perspective segmentation, new tasks like panoramic panoptic segmentation~\cite{prf}, video segmentation~\cite{Panovos}, self-supervised segmentation~\cite{Pano-SfMLearner}, occlusion-aware seamless segmentation~\cite{oass}, and open-vocabulary panoramic segmentation~\cite{ooops} have also been proposed. 

\begin{figure*}[!t]
  \centering
  \includegraphics[width=\linewidth]{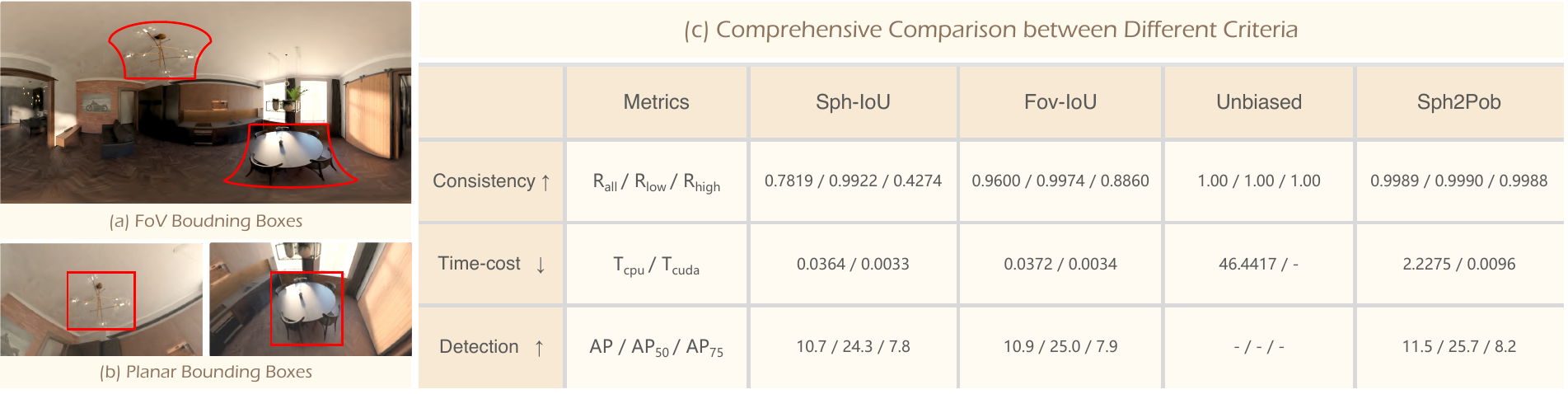}
      \vspace{-15pt}
  \caption{Bounding box representations and evaluation criteria for panoramic object detection. (a) Field of view (FoV) bounding boxes and (b) planar bounding boxes. (c) Metric comparison in consistency, computational cost, and detection accuracy.}
    \label{fig:detection}
    \vspace{-15pt}
\end{figure*}

\subsubsection{Semantic Mapping}

Semantic mapping converts egocentric panoramic inputs into bird’s-eye view (BEV) representations, emphasizing spatial localization of objects rather than pixel-level segmentation. 
Recent works include introducing intermediate projections and distortion-aware indexing (360Mapper~\cite{360bev}); deployment-oriented designs learn direct ERP-to-BEV mappings for efficiency (OneBEV~\cite{onebev}); and fusion-based approaches extend to panoramic–LiDAR integration to mitigate occlusion and FoV limitations in robotics (HumanoidPano~\cite{zhang2025humanoidpano}).

\subsubsection{Object Detection} 

Object detection locates and classifies object instances by predicting bounding boxes and class scores. Unlike semantic segmentation, detection in panoramas faces unique challenges, as conventional rectangular boxes fail to represent object shapes and positions under ERP distortions, particularly near the poles, as illustrated in Fig.~\ref{fig:detection}(a)(b). 
To address this, panoramic detection adopts spherical bounding boxes, angular bounding boxes, or polygon/mask-based annotations that better align with object geometry on the panoramic images. Existing methods can be broadly categorized into three types.

\noindent \textbf{Distortion-Aware Methods} explicitly embed spherical priors into the detection model. 
Approaches include spherical convolutions that define operations on tangent planes or reparameterized kernels to preserve consistency (SphereNet~\cite{spherenet}, SPHCONV~\cite{sphconv}), distortion-aware augmentations with multi-scale kernels or lightweight modules (Curved-Space Faster R-CNN~\cite{curvedspacercnn}, MS-RPN~\cite{msrpn}), and task-specific adaptations such as dynamic token partitioning or deformable context modules (PDAT~\cite{pdat}, PanoGlassNet~\cite{chang2024panoglassnet}).

\noindent \textbf{Projection-Driven Methods} mitigate spherical distortion by decomposing panoramas into multiple distortion-free perspective views, enabling reuse of conventional 2D detectors. Some works, such as Multi-Projection YOLO~\cite{yang2018object} and Rep R-CNN~\cite{reprcnn} refine detection through bounding box adjustment, soft selection, and reprojection-based RoI alignment.

Overall, the above two strategies achieve end-to-end panoramic detection through architectural design, but they overlook the proper representation of spherical bounding boxes, which is a fundamental distinction of panoramic detection.

\noindent \textbf{Redefining Spherical Bounding Boxes and IoU Metrics} builds unbiased spherical representations to better capture object geometry in panoramas. 
Some metrics directly model overlap on the sphere via spherical rectangles or great-circle angles (Sph-IoU~\cite{zhao2020spherical}, FoV-IoU~\cite{foviou}, and Unbiased IoU~\cite{unbiasediou}), eliminating the approximation errors of planar IoU. 
As shown in Fig.~\ref{fig:detection}(c), these spherical metrics achieve higher consistency and more faithful evaluation but often come with increased computational cost. 
In contrast, transformation-based approaches such as Sph2Pob~\cite{sph2pob} map spherical boxes to rotated planar ones, reducing complexity and offering competitive accuracy, though with limited geometric fidelity.

\subsubsection{Visual Tracking}

Visual tracking aims to track the spatial positions and states of objects in video sequences. 
The unique panoramic characteristics from spherical projection lead to structural degradation when directly applying perspective-based algorithms, making it difficult to build consistent spatiotemporal continuity. 
Early efforts enhance object continuity with hardware-driven active vision systems~\cite{jiangtracking}. Cross-modal approaches integrate panoramic images with LiDAR point cloud for geometry-aware detection, association, and bipartite matching (MMPAT~\cite{mmpat}). Multi-camera settings are also extended to panoramic space through pre-fused 3D detections for consistent identity assignment (CC3DT~\cite{cc3dt}). 
To mitigate distortion and boundary issues, spherical representations and metrics such as BFoV, eBFoV, and Sdual are designed to support new 360\textdegree{} benchmarks~\cite{360vot, 360vots}. 
Recent frameworks unify tracking-by-detection and end-to-end paradigms under panoramic designs, enhancing association and representation with trajectory feedback and segmentation cues (Luo et al.~\cite{luo2025omnidirectional}).

\subsubsection{Pose Estimation}

Pose estimation aims to estimate the 6-DoF transformation between images. While geometry-based feature matching methods (e.g., SIFT, ORB) have been widely applied in perspective tasks like SLAM and SfM, panoramic images suffer performance drops in feature detection, matching, and modeling due to their unique characteristics distinct from perspective images.
Recent solutions explore different paradigms: co-visibility-based approaches enhance matching and extend pairwise estimation to multi-view optimization (CoVisPose~\cite{covispose}, Graph-CoVis~\cite{Graph-CoVis}), while self-supervised strategies leverage photometric losses and rotation-only pretraining to handle large viewpoint differences without ground truth poses (PanoPose~\cite{panopose}).

\subsubsection{Saliency Prediction}

Saliency prediction aims to simulate the distribution of human visual attention under spherical viewing conditions. Early traditional methods~\cite{bur2006robot, bogdanova2008visual} rely on low-level features such as intensity, color opponency, and corner responses. With the success of learning-based saliency models on perspective images, transferring them to omnidirectional scenarios remains challenging due to ERP distortions, the scarcity of large-scale 360° benchmarks, and optimization difficulties. Existing approaches can be broadly categorized into three types: 

\noindent \textbf{Projection-Driven Methods} directly project panoramas into Cubemap Projections (CP). Some early works~\cite{monroy2018salnet360, suzuki2018saliency, dai2020dilated, djemai2020extending} adapted conventional 2D saliency models under this paradigm and fused the cubemap-based results into the original panoramas. These methods effectively avoid ERP-specific distortions with well-trained perspective networks, but fail to bridge the feature-level gap. Subsequent research noticed this challenge and proposed methods with integration of both ERP-based global features and CP-based local features.
These approaches~\cite{chao2018salgan360, chen2020salbinet360, dedhia2019saliency} typically employ dual-branch networks or separated pipelines, where predictions from ERP and CP are fused via weighted averaging or spherical-domain optimization.

\noindent \textbf{Spherical Graph Neural Network-based Methods} build graph structures directly on the sphere and apply graph convolutional networks (GCNs) to preserve geometric continuity and spatial adjacency without relying on specific projections~\cite{salgcn, salregcn360, yang2021salgfcn, zhu2021viewing}. They typically sample uniformly on the spherical surface (e.g., geodesic grids or superpixels), where saliency reasoning is performed by propagating features across nodes in geodesic space. 360Spred~\cite{yang2024360spred} further incorporates spherical optical flow and 3D separable graph convolution to jointly capture spatial–temporal saliency patterns.

\noindent \textbf{Behavior Modeling Methods} integrate viewing priors such as equator bias, head–eye motion, or scanpaths, often guided by eye-tracking or viewport trajectories~\cite{de2017look, zhu2018prediction}. For video, behavior-aware approaches further incorporate head-movement prediction, FoV dynamics, and viewport biases via sequence models or multi-task learning~\cite{panosalnet, viewport}, leverage cube-based spatial–temporal designs for geometric continuity~\cite{cubepadding}, simulate scanpaths with reinforcement learning~\cite{dhp}, or fuse global–local cues in dual-stream architectures~\cite{atsal}. 
More recent multimodal frameworks align audio–visual attention for fine-grained saliency~\cite{guo2024instance, cokelek2025spherical}.

\noindent \textbf{Others.} These approaches operate directly on ERP images and introduce novel modules such as attention mechanisms, spherical-specific strategy.
Lightweight solutions improve efficiency and discriminative ability through mobile backbones~\cite{zhu2021lightweight}, dynamic convolutions, or ranking-based attention~\cite{ransp, zhu2021saliency}; spherical U-Net~\cite{zhang2018saliency} defines convolution kernels on spherical crowns to preserve geometric fidelity.

\subsubsection{Layout Detection}

Layout detection aims to recover the structural boundaries of indoor scenes, including walls, floors, and ceilings from panoramic images. However, severe geometric distortions violate traditional assumptions such as planarity and linearity. To address this, existing methods can be broadly categorized into three types:

\noindent \textbf{Projection-Driven Methods} align perspective and panoramic views within a unified spatial framework, reducing inconsistencies across representations. Some approaches include transferring features via dual-branch transformation (DuLa-Net~\cite{dulanet}), integrating cross-view projection with a stereo transformer (PSMNet~\cite{wang2022psmnet}), and converting perspective inputs into ERP for unified processing (uLayout~\cite{ulayout}).

\noindent \textbf{Geometric-Based Methods} adapt to panoramic geometry by converting 2D layouts into 1D horizon sequences for efficiency (HorizonNet~\cite{horizonnet}) and decoupling modeling along orthogonal planes to resolve spatial ambiguities (DOPNet~\cite{dopnet}, MV-DOPNet~\cite{shen2024360}). Building on this principle, C2P-Net~\cite{zhang2025c2p} models components along principal axes and compresses them into 1D representations to enhance spatial reasoning without perspective priors. Differentiable geometric modules further bridge layout and depth (LED$^2$-Net~\cite{wang2021led}, Seg2Reg~\cite{sun2024seg2reg}), while distortion-aware encodings improve spatial consistency (LGT-Net~\cite{lgtnet}). Unified frameworks like HoHoNet~\cite{hohonet} combine sparse 1D and dense 2D predictions in a shared latent space, facilitating multi-task transfer across layout, depth, and semantics.

\noindent \textbf{Structural-Aware Methods} improve the robustness and generalization by integrating geometric priors, ambiguity modeling, and data-efficient strategies. 
Early works adopt classical cues such as lines and vanishing points~\cite{fernandez2018layouts, zhang2014panocontext}, while GPR-Net~\cite{gprnet} leverages learned geometric tokens and multi-view correspondences for layout registration and pose regression. To address annotation ambiguity in existing datasets, dual-branch frameworks like Bi-Layout~\cite{tsai2024no} jointly predict closed and open layout types with cross-attentive refinement. Beyond Manhattan-world assumptions, normal-aware pipelines~\cite{jia20223d} adaptively reconstruct 3D structures under mixed constraints. Meanwhile, data-efficient methods further reduce labeling demands through semi-supervised learning~\cite{tran2021sslayout360} or structure-aware perturbations~\cite{zhang2025semi}, enabling layout estimation in low-resource scenarios.

\subsubsection{Optical Flow Estimation}

Optical flow estimation aims to compute dense motion fields between panoramic video frames. Conventional assumptions such as brightness constancy and spatial smoothness often fail under severe panoramic distortions. Early work like Cubes3D~\cite{apitzsch2018cubes3d} adapts deep optical flow models with custom projection models and synthetic data, highlighting domain gaps between perspective and panoramic settings. Recent methods can be broadly divided into two categories.

\noindent \textbf{Distortion-Aware Methods} adapt convolutional operations to accommodate the geometric distortions for more accurate pixel-wise motion estimation. 
The proposed methods with deformable convolutions (SimpleNet~\cite{xie2019effective}), distortion-aware kernels aligned with spherical geometry (OmniFlowNet~\cite{artizzu2021omniflownet}), angular-aware kernels (LiteFlowNet360~\cite{bhandari2021revisiting}), flow-specific augmentations with cyclic estimation (PanoFlow~\cite{panoflow}), and ortho-driven distortion compensation (PriOr-Flow~\cite{liu2025prior}).

\noindent \textbf{Projection-Driven Methods} integrate motion fields from multiple projection domains. 
The proposed methods utilize a gnomonic-based cubemap and icosahedron fusion, employing off-the-shelf models (Yuan et al.\cite{yuan2021360}), and a joint learning approach for equirectangular, cylindrical, and cubemap flows through a fusion network (Li et al.\cite{li2022deep}).

\subsubsection{Keypoint Matching.} 

Keypoint matching aims to extract repeatable keypoints, compute descriptors on the sphere, and establish correspondences invariant to spherical rotations and robust to ERP distortions.
Traditional methods mitigate distortions by locally approximating the sphere as planar, e.g., SPHORB~\cite{zhao2015sphorb} uses geodesic grids for uniform sampling, and Chuang et al.\cite{chuang2018rectified} project local patches onto tangent planes for description. Learning-based approaches embed spherical geometry into models, such as PanoPoint~\cite {panopoint}, which performs detection/description on ERP images, and SphereGlue~\cite{sphereglue}, which leverages spherical graphs with Chebyshev convolutions. More recently, EDM~\cite{edm} introduces a spherical spatial alignment module with geodesic refinement, enabling coherent yet locally precise dense correspondences.

\subsubsection{Decomposition}

Decomposition separates 360\textdegree{} imagery into interpretable components such as lighting, reflectance, and geometry, enabling realistic understanding and content editing in immersive settings. The proposed methods with stereo-based full-scene illumination and multi-scale learning (Li et al.~\cite{li2021lighting}), physics-driven decomposition via differentiable rendering (PhyIR~\cite{li2022phyir}), and spherical-constrained optimization with intrinsic priors (Xu et al.~\cite{xu2024intrinsic}).

\subsubsection{Lighting Estimation}

Lighting estimation aims to recover high-fidelity environmental illumination, supporting tasks such as inverse rendering, relighting, object insertion, and augmented reality. The goal is typically to create an HDR environment map or a set of spherical harmonics (SH) coefficients that capture the global scene illumination. Early approaches directly infer illumination from full 360\textdegree{} panoramas~\cite{weber2018learning, gkitsas2020deep}, but collecting large-scale panoramic data is impractical. Recent studies instead estimate lighting from limited observations, such as perspective images or object-centric RGB-D views, and can be broadly categorized into two main directions.

\noindent \textbf{Physics-Based Methods} directly regresses compact parameter sets such as sun position, sky turbidity, or SH coefficients, using physically based sky models, autoencoder-learned codes, or CNN predictors. Outdoor-oriented approaches target natural illumination~\cite{hold2017deep, hold2019deep, zhang2019all}, while others address complex indoor lighting with spatial variations~\cite{song2019neural, garon2019fast}. Geometry and reflectance priors, often combined with differentiable rendering layers, provide supervision and support photorealistic relighting.

\noindent \textbf{Panorama Generation Methods} predict full 360\textdegree{} environment maps as an intermediate or final lighting representation, providing more realistic relighting and greater editing flexibility than parameter-only regression. Generative approaches reconstruct HDR panoramas from partial observations using GANs or diffusion models (StyleLight~\cite{wang2022stylelight}, HDRGAN~\cite{somanath2021hdr}, IllumiDiff~\cite{shen2025illumidiff}, Hilliard et al.~\cite{hilliard2025hdr}); physically inspired methods integrate geometric parameterizations or scene cues with editable HDR illumination (EMLight~\cite{zhan2021emlight}, GMLight~\cite{zhan2022gmlight}, Weber et al.\cite{weber2022editable}, EverLight\cite{dastjerdi2023everlight}); and structured pipelines adopt intrinsic decomposition, hierarchical transformers, or latent-diffusion refinement (SOLID-Net~\cite{zhu2021spatially}, SALENet~\cite{zhao2024salenet}, CleAR~\cite{zhao2024clear}). Collectively, these strategies enhance realism, editability, and consistency in panoramic lighting estimation.

\subsubsection{Depth Estimation} 

Depth estimation aims to infer per-pixel scene distances (or disparities) from images, producing dense 3D structural representations. In panoramas, spherical geometry and severe ERP distortions make this task more challenging than in perspective settings. The first learning-based approach, OmniDepth~\cite{zioulis2018omnidepth}, attempts to address this perspective-to-panorama gap by introducing a dedicated dataset and a distortion-aware network, subsequent methods can be broadly categorized into four paradigms.

\noindent \textbf{Distortion-Aware Methods} for depth estimation include distortion-aware convolutions that adapt kernel sampling to non-uniform resolution (Tateno et al.\cite{tateno2018distortion}, ACDNet\cite{zhuang2022acdnet}), Transformer-based architectures with spherical encodings or attention (PanoFormer~\cite{shen2022panoformer}, EGFormer~\cite{yun2023egformer}), diffusion frameworks modeling global structure probabilistically (OmniDiffusion~\cite{mohadikar2025omnidiffusion}), and harmonic-space representations capturing frequency-domain topology (HuSH~\cite{lee2025hush}).

\noindent \textbf{Projection-Driven Methods} for depth estimation include:

\noindent \textit{(a) Multi-Projection Fusion.} Some works include BiFuse~\cite{wang2020bifuse} and UniFuse~\cite{jiang2021unifuse} with dual-branch ERP–cubemap fusion via learnable masks and distortion-aware padding, perspective-view synthesis for local refinement~\cite{peng2023high}, cubemap vision transformers~\cite{bai2024glpanodepth}, and feature-alignment modules~\cite{ai2023hrdfuse}. More advanced designs such as Elite360D/M~\cite{ai2024elite360d, ai2024elite360m} integrate ERP and spherical grids (e.g., ICOSAP) with attention mechanisms and multi-task objectives for improved efficiency and generalization.

\noindent \textit{(b) Projection Transformation.} Instead of fusing views, these methods directly replace ERP with distortion-reduced projections. Tangent patches~\cite{li2022omnifusion, rey2022360monodepth}, icosahedral meshes~\cite{benny2025sphereuformer}, and HEALPix-sampled spheres~\cite{li2023mathcal} enable uniform spatial reasoning and accurate depth regression. 
Further improvements include patch-wise fusion with geometry-aware modules~\cite{li2022omnifusion, mohadikar2024ms360}, spherical transformers~\cite{zhang2025sgformer}, frequency-domain designs such as Gabor-based priors~\cite{shen2024revisiting}, and Cassini-based omnidirectional stereo~\cite{deng2025omnistereo} for geometry-preserving multi-view depth.

\noindent \textbf{Other Methods.} Beyond structure-aware modeling and projection-based strategies, several complementary directions further advance panoramic depth estimation. Slice-based representations partition ERP into gravity-aligned slices or panels to exploit indoor regularities (SliceNet~\cite{pintore2021slicenet}, HoHoNet~\cite{sun2021hohonet}, PanelNet~\cite{yu2023panelnet}); multi-task learning jointly predicts depth with normals or planar boundaries to enforce geometric consistency (Huang et al.\cite{huang2024multi}, Eder et al.\cite{eder2019pano}, Feng et al.\cite{feng2020deep}); self- and weakly supervised methods leverage spherical view synthesis or stereo photometric cues for scalable training on unlabeled panoramas\cite{zioulis2019spherical, wang2023distortion, yun2022improving, zhuang2023spdet, garg2016unsupervised}; and pretraining/foundation adaptation transfers large perspective-based models (e.g., Depth Anything) to panoramas via projection conversion, pseudo-label distillation, and Möbius or equator-aware augmentations~\cite{wang2024depth, cao2025panda}. Recent generalizable frameworks such as Unik3D~\cite{piccinelli2025unik3d} and Depth Any Camera~\cite{guo2025depth} further demonstrate the ability to handle multiple input types, including ERP panoramas.

\subsection{Multi-modal Understanding} 

These works extend panoramic perception beyond vision-only models by integrating complementary modalities such as audio, LiDAR, and text. With the growing adoption of 360\textdegree{} images and videos in VR/AR, this fusion enables richer semantic understanding and human-aligned multimodal perception and generation.

\subsubsection{Audio–Visual Fusion} 

These methods leverage spatial audio to complement panoramic vision with orientation cues and event-specific signals, enabling richer 360\textdegree{} scene understanding beyond vision alone. Unlike perspective settings, panoramic scenarios require spatialized audio over the full sphere with continuity and distortion handling. Recent research spans four representative directions.

\noindent \textit{(a) Spatial Audio Synthesis.} Methods generate ambisonic sound aligned with panoramas from static or mono inputs, using depth cues, geometric simulation, or end-to-end audio–visual networks (Audible Panorama~\cite{huang2019audible}, Li et al.~\cite{li2018scene}, Morgado et al.~\cite{morgado2018self}).

\noindent \textit{(b) Audio–Visual Attention Modeling.} Large-scale eye-tracking datasets and cross-modal saliency frameworks demonstrate how spatial audio guides visual focus under ERP and cubemap projections (AVS-ODV~\cite{zhu2023audio}, PAV-SOD~\cite{zhang2023pav}, PAV-SOR~\cite{guo2024instance}).

\noindent \textit{(c) Semantic Reasoning and Sound Source Localization.} Panoramic tasks require spherical encodings and long-range reasoning for object-level auditory understanding, with methods addressing AVQA, localization, and semantic prediction via spatial encodings, self-supervision, or multi-task learning (Pano-AVQA~\cite{yun2021pano}, Masuyama~\cite{masuyama2020self}, Vasudevan et al.~\cite{vasudevan2020semantic}).

\noindent \textit{(d) Perceptual Quality and 4D Fusion.} Audio-aware models outperform vision-only baselines in panoramic quality assessment~\cite{zhu2023perceptual, fela2022perceptual}, while Sonic4D~\cite{xie2025sonic4d} integrates video-to-audio generation, grounding, and room simulation for realistic 4D interactive experiences.

\subsubsection{Fusion Perception with LiDAR}

These approaches integrate panoramic imagery with sparse LiDAR to enhance robustness in driving, robotics, and mapping. Unlike perspective fusion~\cite{wan2025rapid,yue2025roburcdet} with narrow FoV and rectilinear geometry, panoramic settings require spherical modeling and full-scene alignment. Recent methods explore semantic mapping and planning with panoramic depth cues (SPOMP~\cite{miller2024air}, HumanoidPano~\cite{zhang2025humanoidpano}), indoor reconstruction with fisheye–LiDAR fusion or color-consistent point clouds (Ma et al.\cite{ma2020method}, OmniColor\cite{liu2024omnicolor}), cross-modal saliency and viewpoint-robust localization via spherical attention or equivariant networks (Zhao et al.\cite{zhao2022attention}, Bernreiter et al.\cite{bernreiter2021spherical}), and trajectory estimation or tracking with LiDAR-enhanced optimization (PanoramicVO~\cite{yuan2024panoramic}, MMPAT~\cite{mmpat}).

\subsubsection{Fusion with Text}

Extending vision–language fusion from perspective images to 360\textdegree{} panoramas introduces unique gaps: spherical projections cause severe distortions and wraparound boundaries, while semantics are scattered across the full sphere, making conventional attention and alignment ineffective. Recent methods introduce spatially-aware attention, guided FoV exploration, and spherical encodings to enable effective VQA, captioning, and dense grounding in panoramic contexts, highlighting the importance of geometry-aware fusion for robust panoramic vision–language understanding.

VQA-360~\cite{chou2020visual} pioneers the task with an observe-then-answer strategy and the first panoramic VQA dataset. VIEW-QA~\cite{song2024video} extends this to 360\textdegree{} videos with annotations for assistive scenarios. OmniVQA~\cite{zhang2025towards} and OSR-Bench~\cite{dongfang2025multimodal} further emphasize spatial reasoning and multimodal grounding, evaluating MLLMs via structured policy learning and cognitive map–based protocols. Most recently, Dense360~\cite{zhou2025dense360} introduces dense captioning and grounding for ERP images using an ERP-RoPE encoding and large-scale datasets, enabling spatially consistent language grounding in panoramic contexts.

\begin{figure*}[t]
\centering
\includegraphics[width=1\linewidth]{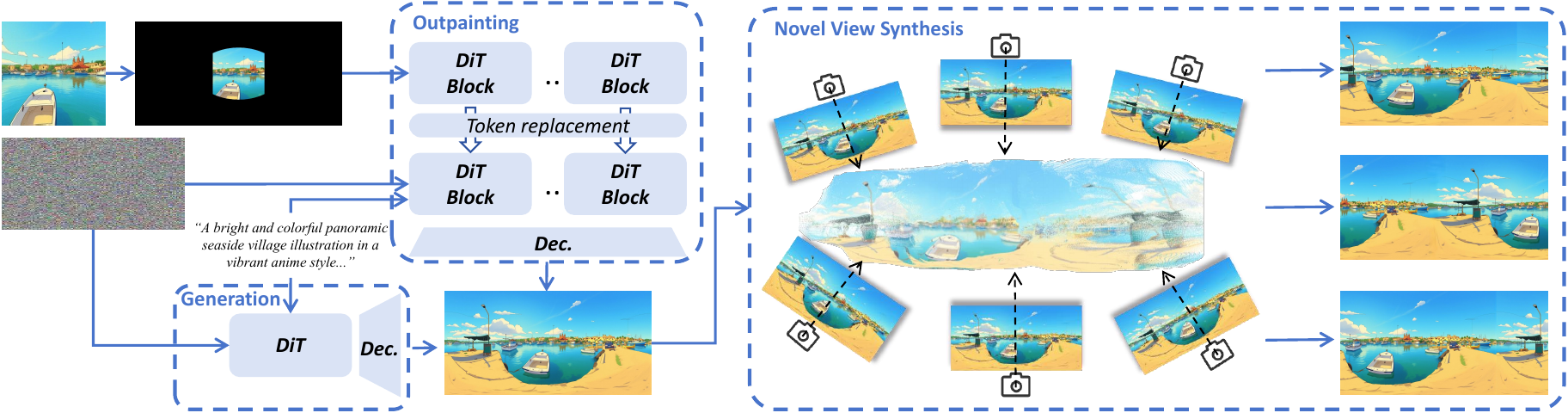}
\caption{Overview of a panoramic world modeling pipeline, which integrates text-guided generation, image completion, and novel view synthesis. These stages also correspond to three representative tasks, demonstrating how generative methods collectively advance holistic scene modeling.}
\vspace{-10pt}
\label{fig:generation}
\end{figure*}

\subsection{Generative-modeling-based Tasks}

With the rising interest in world models, an emerging approach leverages the inherent geometric consistency of panoramas as a foundation for generation. Recent advances such as Matrix-3D~\cite{yang2025matrix} and HunyuanWorld 1.0~\cite{team2025hunyuanworld} exemplify this trend, pushing panoramic generation beyond single-scene synthesis toward omnidirectional, explorable 3D worlds. 
Within this paradigm, as illustrated in Fig. \ref{fig:generation}, text-guided generation introduces semantic controllability through multimodal conditioning, image completion recovers unobserved content in missing or outpainted regions, and novel view synthesis extends NeRF and 3DGS frameworks to panoramic settings, enabling faithful viewpoint expansion under spherical geometry.
Together, these components drive the development of world modeling, shifting 360\textdegree{} vision from reconstruction of observations into the generative construction of rich, interactive, and semantically grounded virtual worlds.

\subsubsection{Text-guided Generation}

Text-guided generation aims to synthesize panoramic images or videos from textual descriptions, providing semantic controllability for 360\textdegree{} content creation.
Diffusion models have driven substantial advances in text-to-image/video (T2I/V) synthesis for perspective images and videos, enabling high-quality and controllable generation~\cite{yang2025unified,qi2024unigs}. Extending these models to 360\textdegree{} panoramas is nontrivial due to spherical topology, which induces geometric distortion and discontinuities across boundaries. Geometry-aware generation is therefore essential, with clear benefits for downstream tasks requiring precise spatial understanding. Early attempts like Text2Light~\cite{chen2022text2light} remain widely used benchmarks but do not explicitly enforce spherical consistency or address panorama-specific structural and semantic issues, and thus serve as transitional rather than definitive solutions to the domain gap.

\noindent \textbf{Distortion-Aware Methods} explicitly adapt architectures to spherical geometry and ERP distortions for panoramic generation. Representative designs include spherical convolutions (SMConv, SMGD~\cite{sun2025spherical}) and SphereDiffusion~\cite{wu2024spherediffusion} for native spherical modeling, pixel-wise reweighting (PanoWAN~\cite{xia2025panowan}) and projection-specific denoising (DynamicScaler~\cite{liu2025dynamicscaler}) for distortion-aware optimization, as well as transformer backbones with spherical encodings (PanoDiT~\cite{zhang2025panodit}) to capture long-range dependencies while preserving geometric fidelity.

\noindent \textbf{Projection-Driven Methods} for panoramic generation include tangent-plane decomposition (TanDiT~\cite{ccapuk2025tandit}), dual-branch architectures that combine perspective and panoramic contexts (Zhang et al.~\cite{zhang2024taming}, TiP4GEN~\cite{xing2025tip4gen}), RGB-Depth cube diffusion
framework (DreamCube~\cite{huang2025dreamcube}), and hybrid spherical–planar modeling (SphereDiff~\cite{park2025spherediff}). Other approaches build features from overlapping perspective views or predefined tangent directions (VideoPanda~\cite{xie2025videopanda}, ViewPoint~\cite{fang2025panoramic}), achieving localized consistency while maintaining global coherence.

\noindent \textbf{Continuity Modeling Methods} enhances wraparound continuity in 360\textdegree{} panoramas through inference-time strategies without altering base architectures. Training-free approaches include circular blending (Diffusion360~\cite{feng2023diffusion360}), dual pre-denoising at borders (Wang et al.\cite{wang2024customizing}), circular padding (Wang et al.\cite{wang2025conditional}, 360DVD\cite{wang2024360dvd}), and iterative warping with bidirectional guidance (PanoFree~\cite{liu2024panofree}), among which PanoFree stands out as a plug-and-play solution balancing efficiency and compatibility.

\noindent \textbf{Other Methods} explore complementary directions such as coarse-to-fine temporal modules for panoramic video generation (VidPanos~\cite{ma2024vidpanos}) and user-controllable approaches like OmniDrag~\cite{li2024omnidrag}, which explicitly model spherical motion and enable trajectory-based editing through temporally aware architectures and motion-diverse datasets.

\subsubsection{Image Completion}

Image completion, including both inpainting (filling missing regions) and outpainting (extending beyond boundaries), has achieved notable success in perspective images using autoencoders, GANs, and diffusion models. However, extending to panoramas remains challenging due to ERP-induced distortions around poles and seams, the difficulty of generating plausible content in large missing regions, and poor handling of edge- and pole-specific structures such as sky, ground, and stitching artifacts. To address these issues, recent research explores two main directions.

\noindent \textbf{Distortion-Aware Methods} for 360\textdegree{} completion include Dream360~\cite{ai2024dream360}, which builds spherical latent spaces to reduce planar bias; PanoDecouple~\cite{zheng2025panorama}, which separates distortion guidance and content completion with distortion maps and Distort-CLIP loss; and 2S-ODIS~\cite{nakata20242s}, which employs a two-stage VQGAN + NFoV refinement pipeline to balance global layout and local detail. Together, these methods improve structural alignment and perceptual fidelity in panoramic completion.

\noindent \textbf{Continuity Modeling Methods} enforce circular consistency in panoramic completion. Cylin-Painting~\cite{liao2023cylin} preserves circular continuity with cylinder-style convolutions, PanoDiff~\cite{wang2023360} and PanoDiffusion~\cite{wu2023panodiffusion} achieve rotation equivariance and wraparound consistency with distortion–content decoupling and RGB-D cues, while Akimoto et al.~\cite{akimoto2022diverse} enhance texture fidelity and semantic continuity through a dual-network design with circular inference. Collectively, these methods highlight the importance of explicitly modeling boundary continuity to achieve seamless panoramic completion.

\noindent \textbf{Other Methods} explore complementary directions beyond distortion correction and continuity modeling, including semantic conditioning (ImmerseGAN~\cite{dastjerdi2022guided}), multi-modal autoregressive generation with NFoV, text, and geometry cues (AOG-Net~\cite{lu2024autoregressive}), and structure-aware RGB-D disentanglement with new evaluation metrics (BIPS~\cite{oh2022bips}).

\subsubsection{Novel View Synthesis}

Novel View Synthesis (NVS) for 360\textdegree{} panoramas aims to generate unseen viewpoints from limited observations.
While perspective-based methods such as NeRF~\cite{mildenhall2021nerf} and 3D Gaussian Splatting (3DGS)~\cite{3dgs} have achieved remarkable progress, directly extending them to panoramic faces faces challenges including spherical distortions, wide-FoV occlusions, and inefficiencies in Cartesian sampling. Recent approaches address these issues by incorporating spherical representations, depth priors, and soft occlusion handling into NeRF/3DGS frameworks, and further explore hybrid architectures and AIGC-driven pipelines for efficient, semantically consistent panoramic scene generation.

\noindent \textbf{NeRF-based Methods} adapt neural radiance fields to 360\textdegree{} panoramas by addressing spherical distortion, wide-baseline occlusions, and inefficient Cartesian sampling. Early efforts reformulate ray sampling and feature aggregation in spherical coordinates to improve efficiency and geometric consistency (EgoNeRF~\cite{choi2023balanced}, PanoGRF~\cite{chen2023panogrf}); later studies introduce novel camera models (OmniNeRF~\cite{gu2022omni}), local radiance field partitioning (OmniLocalRF~\cite{choi2024omnilocalrf}, 360Roam~\cite{huang2022360roam}), and semantic or depth priors for sparse-view or monocular inputs (360FusionNeRF~\cite{kulkarni2023360fusionnerf}, Perf~\cite{wang2024perf}) to enhance scene understanding from sparse or monocular inputs. while latest studies incorporate HDR estimation~\cite{gera2022casual} or inpainting-based augmentation to mitigate data scarcity.

\noindent \textbf{3DGS-based Methods} extend 3D Gaussian Splatting to panoramic view synthesis by addressing projection mismatch, sparse inputs, and ERP distortion through geometry-aware projection, sampling, and rasterization. To align with spherical geometry, differentiable splatting on tangent planes or Fibonacci lattices reduces pole distortion and improves ERP rendering (360-GS~\cite{bai2024360}, ODGS~\cite{lee2024odgs}, PanSplat~\cite{zhang2025pansplat}); OmniGS~\cite{li2024omnigs} replaces cubemap approximations with differentiable projection models for stronger generalization. Dual-projection encoders and Yin–Yang decompositions enhance Gaussian parameter prediction (Splatter-360~\cite{chen2025splatter}, OmniSplat~\cite{lee2025omnisplat}), while layout priors, boundary optimization, hierarchical cost volumes, RoPE rolling, and distortion-aware losses, dual-Fisheye distortion modeling improve robustness in sparse or low-texture indoor scenes (360-GS~\cite{bai2024360}, TPGS~\cite{shen2025you}, PanoSplatt3R~\cite{ren2025panosplatt3r}, ErpGS~\cite{ito2025erpgs}, Seam360GS~\cite{shin2025seam360gs}). 
Finally, OB3D~\cite{ito2025ob3d} introduces a high-fidelity omnidirectional dataset with diverse trajectories, enabling rigorous benchmarking and advancing panoramic 3DGS research.

\noindent \textbf{Other Representations} move beyond volumetric NeRFs and point-based 3DGS by adopting layered images (MSI/LDI) or spherical meshes for geometry-aware warping and differentiable compositing. Specifically, SOMSI~\cite{habtegebrial2022somsi} follows the layered-image paradigm with a soft-occlusion MSI, achieving high quality with few layers and fast feedforward synthesis. OmniSyn~\cite{li2022omnisyn} follows the spherical-mesh paradigm by pairing 360\textdegree{} depth and a spherical cost volume with a differentiable 360\textdegree{} mesh renderer to handle wide baselines and occlusions. Similarly, Casual 6-DoF~\cite{chen2022casual} enhances the spherical-mesh approach by integrating panoramic depths into a lightweight mesh, allowing for real-time, VR-ready 6-DoF navigation from casually captured panoramas.

\subsubsection{Applications.} 

Applications of panoramic generative models build on the spatial consistency and completeness of 360\textdegree{} vision, extending text-driven control to diverse modalities and downstream tasks. Representative directions include constrained scene generation for bounded single-scene synthesis, unconstrained world generation for open-ended multi-trajectory environments, and vision–language navigation (VLN) for embodied interaction. Together, these applications adapt panoramic generation for spatial reasoning, immersive interaction, and controllable environments.

\noindent \textbf{Constrained Scene Generation.} leverages text-to-360\textdegree{} panoramas as holistic spatial priors for constructing immersive 3D and 4D environments. SceneDreamer360~\cite{li2024scenedreamer360} and DreamScene360~\cite{zhou2024dreamscene360} synthesize high-quality panoramas with depth estimation to reconstruct 3D Gaussian Splatting scenes, ensuring high-resolution rendering and spatial fidelity. Similarly, LayerPano3D~\cite{yang2024layerpano3d} lifts layered decompositions into Gaussian fields for progressive optimization, while HoloDreamer~\cite{zhou2024holodreamer} employs a multi-stage depth–geometry pipeline for view-consistent 3D scenes. Extending to dynamics, HoloTime~\cite{zhou2025holotime} generates ERP videos for re-renderable 4D reconstruction, and 4K4DGen~\cite{li20244k4dgen} decomposes panoramas into perspective views and recovers spherical geometry with 4D Gaussian fields.

\noindent \textbf{Unconstrained World Generation} extends panoramic generation from single-scene synthesis to open-ended, explorable 3D worlds. Matrix-3D~\cite{yang2025matrix} employs trajectory-guided panoramic video diffusion conditioned on mesh renders for precise camera control, lifting generated 360\textdegree{} videos into navigable 3D via either fast panorama reconstruction or high-fidelity 3DGS pipelines. Complementarily, HunyuanWorld 1.0~\cite{team2025hunyuanworld} builds interactive worlds from text or a single image through semantic layering, layer-aligned depth, and mesh-based reconstruction, combining foreground asset warping, HDR sky modeling, and world-consistent video diffusion for long-range exploration.

\noindent \textbf{Vision–Language Navigation (VLN)} leverages text-to-360\textdegree{} generation to provide panoramic context and semantically consistent observations for instruction following. PANOGEN~\cite{li2023panogen} and PANOGEN++~\cite{wang2025panogen++} generate realistic panoramas from room descriptions with semantic layouts and orientation-aware object placement, enriching VLN training. VLN-RAM~\cite{wei2025unseen} expands unseen coverage by rewriting scenes and instructions with object-aware augmentation, while Any2Omni~\cite{yang2025omni} introduces a large-scale dataset and a spatially consistent Transformer (Omni$^2$) for panoramic generation and editing.

\begin{figure*}[t]
\centering
\includegraphics[width=1\linewidth]{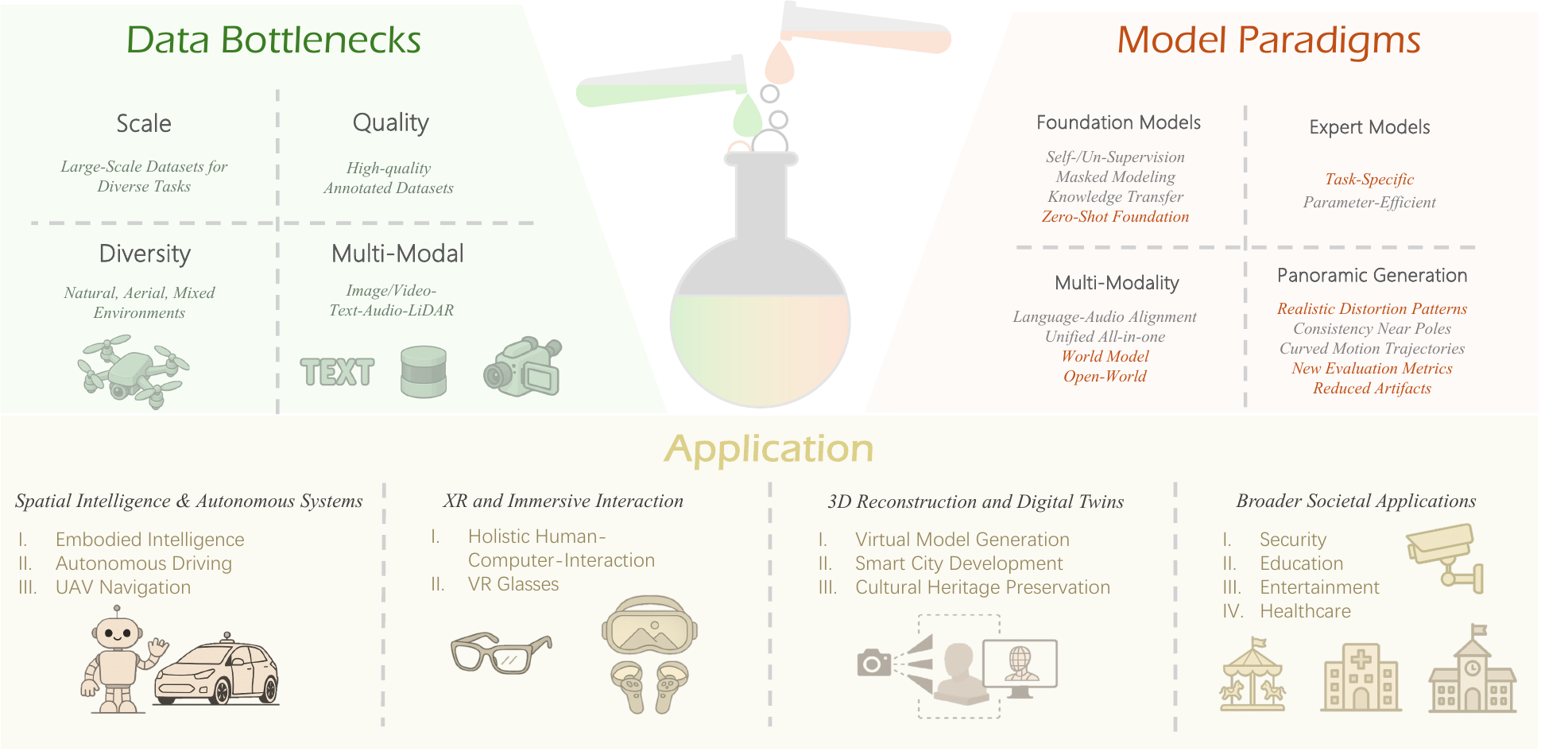}
\vspace{-15pt}
\caption{Summary of future directions in panoramic vision: (1) \textbf{Data Bottlenecks}, emphasizing challenges in scale, diversity, quality, and multi-modality; (2) \textbf{Model Paradigms}, covering foundation models, expert models, multi-modal integration, and models for panoramic generation; and (3) \textbf{Applications}, including spatial intelligence and autonomous systems, XR and immersive interaction, 3D reconstruction and digital twins, and broader societal applications such as security, education, entertainment, and healthcare.}
\vspace{-15pt}
\label{fig:future}
\end{figure*}

\section{Challenges and Future Works}
\label{sec:future}
Despite rapid progress, panoramic vision still faces fundamental challenges that limit its scalability, robustness, and deployment in real-world scenarios. Based on the analysis of existing methods, we highlight future research opportunities from three complementary perspectives: data, models, and applications.

\subsection{Data Bottlenecks} 

Compared with perspective vision, a major bottleneck for panoramic vision is data scarcity. Existing datasets (a comparative summary with perspective datasets is provided in supplementary material) remain limited in scale, diversity, quality, and modality, thereby constraining model generalization and hindering fair benchmarking.

\noindent \textbf{Scale.} Large-scale, standardized 360\textdegree{} image and video datasets across diverse tasks and scenarios remain scarce, constraining model training and reproducible evaluation.

\noindent \textbf{Diversity.} Most existing datasets concentrate on indoor or urban settings, with limited coverage of natural, aerial, or mixed environments, thus seriously constraining progress towards open-world generalization.

\noindent \textbf{Quality.} High-quality annotated panoramic datasets for tasks such as depth estimation, segmentation, detection, tracking, and mapping remain limited, particularly with fine-grained labels in real-world scenarios.

\noindent \textbf{Multi-Modality.} Panoramic image–text and video–text resources remain limited, restricting advances in language-guided generation, VQA, and cross-modal reasoning.

Overall, future progress depends on constructing large-scale, foundational, standardized, multi-modal datasets to enhance generalization, comparability, and performance across various tasks.

\subsection{Model Paradigms} Progress in panoramic vision is closely tied to advances in model design, ranging from general foundation frameworks to task-specific expert architectures, and further extending to multimodal and generative paradigms. 
A major challenge is to develop models that can effectively adapt to panorama-specific representations while progressing toward three key goals: (1) strong ability for generalization and zero-shot transfer, (2) unified all-in-one architectures that jointly support multiple tasks, and (3) world modeling techniques capable of open-world understanding and scene generation.

\noindent \textbf{Foundation Models.} Training paradigms such as self-/un-supervision, contrastive learning, and masked modeling require adaptation to 360\textdegree{} data. An important direction is to transfer knowledge from foundational perspective models to panoramic domains, thereby reducing the domain gap and improving efficiency. Future foundation models should emphasize on zero-shot robustness, ensuring reliable performance in novel panoramic environments under limited supervision.

\noindent \textbf{Expert Models.} In addition to general-purpose foundation models, task-specific expert models remain essential. On the one hand, incorporating panoramic characteristics into task-specific architecture designs can lead to better efficiency and accuracy. On the other hand, for tasks such as detection, segmentation, depth estimation, and temporal analysis, integrating pre-trained base networks with parameter-efficient expert modules can further boost performance while preserving generalization.

\noindent \textbf{Multi-Modality and Panoramic Generation.} Existing multi-modal frameworks still struggle to handle panoramic-specific properties such as spatial continuity and distortion distribution. Future developments may involve incorporating panoramic priors into architectures that better align vision, language, and audio. Detailed promising directions include: (1) panorama–language alignment for improved grounding, (2) unified frameworks that integrate generation and understanding, (3) world models for continuous and interactive panoramic scene synthesis, and (4) open-world adaptation to unseen semantics. 

While understanding and generation are inherently coupled in multi-modal panoramic systems, advancing panoramic generation remains equally important. Key challenges include: (i) developing specialized evaluation metrics, (ii) preserving realistic distortion patterns, (iii) ensuring consistency near poles, and (iv) modeling curved motion trajectories distinct from those in perspective video. In addition, panoramic video generation poses further difficulties in maintaining spatiotemporal coherence.

\subsection{Application}

\noindent \textbf{Spatial Intelligence and Autonomous Systems.} 
360\textdegree{} panoramic vision provides a complete environmental context, which inherently aligns with the demands of spatial intelligence. 
By eliminating blind spots and enhancing global perception, it enables robust scene understanding and decision-making, which is particularly crucial for embodied intelligence, autonomous driving, and UAV navigation, where comprehensive situational awareness directly supports safety and reliability. 

\noindent \textbf{XR and Immersive Interaction.} 
From panoramic recording to high-resolution content generation, 360\textdegree{} vision forms the cornerstone of extended reality (XR). Future directions include integrating spatial audio, haptic feedback, and other multi-sensory modalities to create a holistic, immersive interaction paradigm. Moreover, supporting rich and perceptually aligned human-computer interaction across the human senses, together with lightweight deployment on portable devices such as VR/AR glasses, will drive practical adoption and daily usage. 

\noindent \textbf{3D Reconstruction and Digital Twins.} 
Panoramic imaging captures holistic scene information, enabling the complete reconstruction of 3D environments and the creation of digital twins. 
Applications range from 3D mapping and spatial digital archiving to virtual model generation, supporting fields such as smart city development and cultural heritage preservation.

\noindent \textbf{Broader Societal Applications.} 
Beyond the above technical directions, panoramic vision also holds broad prospects for practical use in various domains. 
It can enhance security and surveillance through full-scene monitoring with fewer blind spots, enrich education and training via immersive content delivery, enable new forms of entertainment and media through high-fidelity 360\textdegree{} capture and generation, and support healthcare with XR-assisted telemedicine and rehabilitation. 
These examples highlight its transformative potential across diverse industries. 

\section{Conclusion}
This survey provides a comprehensive overview of panoramic vision, aiming to bridge the gaps between panoramic and perspective representations. First, we analyze panoramic imaging systems and projection models, which reveal the unique geometric characteristics underlying the fundamental gaps between panoramic and perspective representations: geometric distortion, non-uniform spatial sampling, and boundary continuity.
Next, we conduct both cross-method and cross-task analysis across more than 20 representative tasks, synthesizing common strategies while highlighting their advantages, limitations, and applicability. 
Finally, we outline several future directions, including building larger and more diverse datasets, developing foundational, multimodal, and generation models, and extending to broader downstream applications such as embodied intelligence, autonomous driving, and immersive media.
Overall, this survey serves as both a comprehensive reference and a forward-looking guide for the continued development of panoramic vision.

\section{Acknowledgment}
We would like to express our sincere gratitude to Yunning Peng, Haoran Feng, Shi Luo, Ruihua Lu for their valuable contributions that greatly improved the quality of this paper. We also gratefully acknowledge the generous support of Antigravity Team and Insta360 Research Team, whose assistance in various aspects made this work possible.

{\small
\bibliographystyle{IEEEtran}
\bibliography{e}
}

\clearpage 
\appendix

This supplementary material document provides a detailed introduction of some common projection formats, a comparison of perspective and panoramic datasets, and task-specific future directions.
The supplementary material is organized as follows:
\begin{enumerate}[label=\arabic*.]
    \item Projection Formats.
    \item Comparison of Perspective and Panoramic Datasets.
    \item Future Work by Task.
\end{enumerate}

\section*{Projection Formats}
Here, we provide a more detailed summary of representative projections commonly adopted in panoramic vision beyond the main text, with a focus on their characteristics and limitations.

\noindent \textbf{Spherical Projection.} A 360$^\circ$ camera can be modeled as a pinhole at the center of a unit sphere, projecting all visible 3D points onto its surface without requiring traditional intrinsic parameters~\cite{krolla2014spherical, da20243d, li2005spherical}. 
A 3D point in the world Cartesian coordinate system
$P=[x, y, z]^T$ is first transformed into spherical coordinates $(\rho, \theta, \phi)$, where $\rho=\sqrt{x^2+y^2+z^2}$, $\theta=\arccos(z/\rho)$, and $\phi=\arctan2(y,x)$.
By normalizing with respect to $\rho$, the point is mapped onto the unit sphere, yielding the unit vector $p = (\sin\theta\cos\phi, \ \sin\theta\sin\phi, \ \cos\theta) = (x', y', z')$. The unit sphere is centered at the origin of the world coordinate system.
This spherical projection provides a unified, distortion-free representation of all viewing directions, serving as a fundamental basis for analyzing panoramic images and videos.

\noindent \textbf{Equirectangular Projection (ERP).} As the most common format for 360-degree panoramas, ERP maps spherical coordinates $(\phi, \theta)$ directly and uniformly to a 2D image plane: $\phi$ (longitude) to the horizontal axis and $\theta$ (latitude) to the vertical axis, with sampling intervals $\Delta\phi = 2\pi/w$ and $\Delta\theta = \pi/h$ for image width $w$ and height $h$ (typically $w = 2h$). 
A pixel $(u, v)$ corresponds to $(\phi, \theta) = (u \cdot \Delta\phi - \pi, \ v \cdot \Delta\theta)$. 
Such a simple bijective mapping is analogous to the transformation of the Earth’s spherical surface into a world map, making ERP efficient for rendering, editing, and training vision models. 
Unless specified otherwise, we use ERP as the default projection in this survey.

\noindent \textbf{Cubemap Projection (CMP).} The CMP is a widely used alternative to ERP that alleviates geometric distortions, particularly the stretching near the poles. 
It projects spherical content onto the six faces of a cube, each covering a $90^\circ \times 90^\circ$ field of view. 
Each cube is centered at the camera origin, with faces oriented toward the front, back, left, right, top, and bottom. 
For a 3D point $p = [x', y', z']^T$ on the unit sphere, the face index is determined by the dominant coordinate axis, and the point is projected onto that face via perspective projection. For example, if $P$ maps to the front face, the 2D coordinates are
\[
u = \frac{x}{|z|}, \quad v = \frac{y}{|z|}.
\]
Each face is rendered as a $w \times w$ square, producing a complete 6 unfolded layouts. 
Overall, the CMP provides a distortion-reduced, face-based representation well-suited for panoramic rendering and processing.

\begin{figure}[t]
    \centering
    \includegraphics[width=0.8\linewidth]{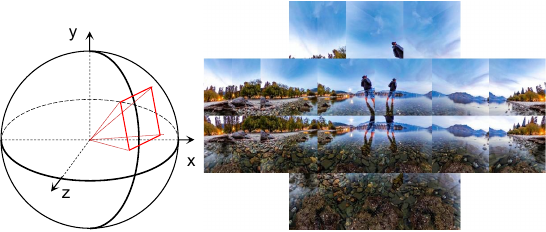}
        \vspace{-5pt}
    \caption{Tangent projection (TP) maps spherical points onto tangent planes via gnomonic projection, preserving local geometry and enabling perspective model reuse.}
    \label{fig:projection_3}
    \vspace{-10pt}
\end{figure}

\noindent \textbf{Tangent Projection (TP).} To adapt perspective models for high-resolution spherical data,~\cite{eder2020tangent} proposes TP, which renders an omnidirectional image as local planar grids tangent to the faces of a subdivided icosahedron. 
As illustrated in Fig.~\ref{fig:projection_3}, each patch is obtained from a gnomonic projection~\cite{karigiannis2020introduction}, mapping a point $P_s$ in the sphere to a tangent plane centered at $P_c$.
A point in the unit sphere with spherical coordinates $(\theta, \phi)$ is projected directly onto a tangent plane centered at $(\theta_c, \phi_c)$, yielding plane coordinates
$(u_t, v_t)$:
\[
\begin{aligned}
u_t &= \frac{\cos\phi \,\sin(\theta-\theta_c)}{\cos c}, \\
v_t &= \frac{\cos\phi_c \,\sin\phi - \sin\phi_c \,\cos\phi \,\cos(\theta-\theta_c)}{\cos c},
\end{aligned}
\]
where $c$ is the central angle between $(\theta, \phi)$ and the tangent point $(\theta_c, \phi_c)$.
$\cos c = \sin\phi_c \sin\phi + \cos\phi_c \cos\phi \cos(\theta-\theta_c)$. 
The inverse mapping follows the same geometric principle, enabling one-to-one correspondence between spherical coordinates and tangent patches~\cite{li2022omnifusion}. 
Overall, TP preserves local geometric fidelity and reduces distortion compared to ERP. 
It also allows the reuse of standard perspective vision models on spherical imagery.

\noindent \textbf{Polyhedron Projection (PP).} To reduce distortions from mapping spherical images to planar formats while preserving directional continuity, PP~\cite{lee2019spherephd} approximates the sphere with a subdivided polyhedron, such as an octahedron or icosahedron. 
Each face of the base polyhedron can be recursively divided into four smaller faces, increasing resolution and reducing distortion~\cite{lee2019spherephd}. 
For instance, an icosahedron at level $l$ yields $20 \times 4^l$ triangular faces, providing a near-uniform sphere sampling.

\noindent \textbf{Panini Projection.} The Panini projection mitigates the strong distortions of rectilinear projection at wide fields of view (typically $>70^\circ$) by preserving vertical and radial lines while non-linearly compressing the horizontal field. 
It maintains a strong central perspective without exaggerating objects near the periphery. 
Specifically, the mapping consists of projecting a unit spherical point $(\phi, \theta)$ to a cylinder, followed by a rectilinear projection from a variable center:
\[
S = \frac{d + 1}{d + \cos{\phi}}, \quad
h = S \cdot \sin{\phi}, \quad
v = S \cdot \tan{\theta},
\]
where $d \geq 0$ controls horizontal compression. $d=0$ yields a rectilinear projection, $d=1$ produces a cylindrical stereographic projection, and $d \to \infty$ approximates the orthographic projection. 
This flexibility enables a smooth trade-off between central magnification and edge compression.

\noindent \textbf{Other projections.} Apart from the aforementioned popular projection formats, there are several other formats supported by the 360Lib software package for coding and processing~\cite{uyen2020subjective}. 
They include adjusted equal-area projection (AEP), truncated square pyramid projection (TSP), adjusted cubemap (ACP), rotated sphere projection (RSP), equatorial cylindrical projection (ECP), equiangular cubemap (EAC), and hybrid equiangular cubemap (HEC). 
Especially, as a map-based projection, AEP adaptively decreases the sampling rate in vertical coordinates and avoids the over-sampling problem in ERP. 
Equi-Angular Cubemap (EAC) projection~\cite{chen2018recent} maintains spatial sampling rates for different sampling locations on the faces of the cubes to alleviate distortions in CP.

\begin{figure}[t]
    \centering
    \includegraphics[width=0.8\linewidth]{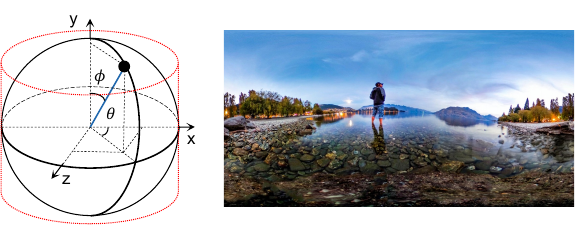}
        \vspace{-5pt}
    \caption{Equirectangular projection (ERP) mapping spherical coordinates ($\phi$, $\theta$) to image pixels ($u$, $v$), analogous to flattening the Earth onto a world map.}
    \label{fig:projection_1}
    \vspace{-10pt}
\end{figure}

\begin{figure}[t]
    \centering
    \includegraphics[width=0.8\linewidth]{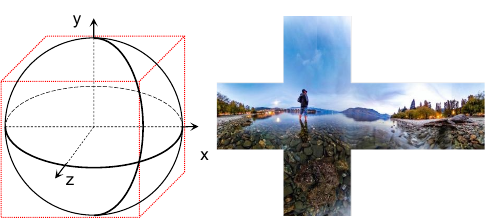}
        \vspace{-5pt}
    \caption{Cubemap projection (CMP) mapping spherical content onto six cube faces via perspective projection, reducing polar distortions compared with ERP.}
    \label{fig:projection_2}
    \vspace{-10pt}
\end{figure}

\begin{figure}[t]
    \centering
    \includegraphics[width=0.8\linewidth, keepaspectratio]{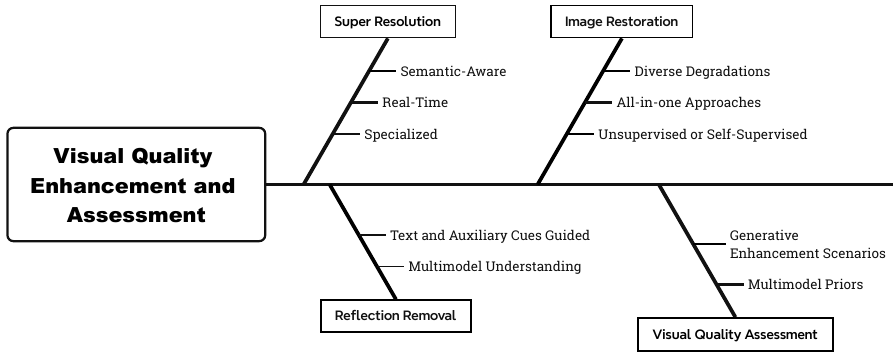}
        \vspace{-5pt}
    \caption{Summary of Future Directions in Visual Quality Enhancement and Assessment.}
    \label{fig:future_1}
    \vspace{-15pt}
\end{figure}

\begin{table*}[t]
\centering
\fontsize{9}{10}\selectfont
\setlength{\tabcolsep}{2.8mm} 
\renewcommand{\arraystretch}{1.15}
\caption{Comparison of Perspective and Panoramic Datasets across Multiple Tasks}
\label{tab:persp_pano_datasets}
\begin{adjustbox}{max width=\textwidth} 
\begin{tabularx}{\textwidth}{
  >{\centering\arraybackslash}p{2.6cm}  
  c c c c c c
}
\toprule
\textbf{Task} & \textbf{Perspective Dataset} & \textbf{Data Size} & \textbf{Source} & \textbf{Panoramic Dataset} & \textbf{Data Size} & \textbf{Source} \\
\midrule
\multirow{4}{*}{\textbf{Generation}}
  & RTMV \cite{tremblay2022rtmv} & 300k & Synthetic & 360SP \cite{chang2018generating} & 15{,}730 & Real \\
  & SPARF \cite{hamdi2023sparf} & 17M  & Synthetic & HDR360-UHD \cite{chen2022text2light} & 4{,}392 & Real \\
  & ACC\mbox{-}NVS1 \cite{sugg2025accenture} & 148k & Real & Laval Indoor HDR \cite{gardner2017learning} & 2{,}100 & Real \\
  & Para\mbox{-}Lane \cite{ni2025lane} & 80k & Real & Laval Outdoor HDR \cite{hold2019deep} & 205 & Real \\
\midrule
\multirow{3}{*}{\makecell{\textbf{Super-}\\\textbf{Resolution}}}
  & GameIR\mbox{-}SR \cite{zhou2024gameir} & 19{,}200 & Synthetic & ODI\mbox{-}SR \cite{deng2021lau} & 1{,}000 & Real \\
  & GameIR\mbox{-}NVS \cite{zhou2024gameir} & 57{,}600 & Synthetic & Flickr360 \cite{cao2023ntire} & 3{,}150 & Real \\
  & ImageParis \cite{joze2020imagepairs} & 11{,}421 & Real & ODV360 \cite{cao2023ntire} & 250 & Real \\
\midrule
\multirow{5}{*}{\textbf{Object Detection}}
  & MS COCO \cite{lin2014microsoft} & 328k & Real & 360\mbox{-}Indoor \cite{chou2020360} & 3{,}000 & Real \\
  & Pascal VOC \cite{everingham2010pascal} & 21k & Real & ERA \cite{yang2018object} & 903 & Real \\
  & Open Images V7 \cite{OpenImages} & 1.9M & Real & OVS \cite{yu2019grid} & 600 & Real \\
  & Objects365 \cite{Shao2019Objects365} & 600k & Real & PANDORA \cite{xu2022pandora} & 3{,}000 & Real \\
  & BDD100K \cite{bdd100k} & 100k & Real & COCO\mbox{-}Men \cite{zhao2020spherical} & 7{,}000 & Synthetic \\
\midrule
\multirow{2}{*}{\textbf{Segmentation}}
  & LVIS \cite{gupta2019lvis} & 164k & Real & PASS \cite{pass} & 1{,}050 & Real \\
  & Open Images V4 \cite{kuznetsova2020open} & 9.2M & Real & WildPASS \cite{yang2021capturing} & 2{,}500 & Real \\
\midrule
\multirow{2}{*}{\textbf{Depth Estimation}}
  & ScanNet \cite{dai2017scannet} & 5M & Synthetic & Stanford2D3D \cite{armeni2017joint} & 1{,}413 & Real \\
  & SDCD \cite{li2024synthetic} & 930k & Real & Deep360 \cite{li2022mode} & 2{,}100 & Real \\
\midrule
\multirow{2}{*}{\textbf{Saliency Prediction}}
  & SALICON \cite{jiang2015salicon} & 10k & Real & Salient360! \cite{rai2017dataset} & 85 & Real \\
  & CAT2000 \cite{borji2015cat2000} & 4k & Real & PVS\mbox{-}HM \cite{xu2021pvs} & 76 & Real \\
\midrule
\multirow{1}{*}{\textbf{Room Layout}}
  & LSUN Room Layout \cite{lin2018layoutestimation} & 5{,}394 & Real & Pano3DLayout \cite{pintore2021deep3dlayout} & 107 & Synthetic \\
\bottomrule
\end{tabularx}
\end{adjustbox}
\end{table*}

\section*{Comparison of Perspective and Panoramic Datasets.}

Table~\ref{tab:persp_pano_datasets} shows a striking imbalance phenomenon between perspective and panoramic datasets across a wide range of tasks. 
For perspective vision, large-scale datasets such as Open Images (9.2M images), Objects365 (600k images), and ScanNet (5M frames) have provided abundant data to train powerful foundation models. 
By contrast, panoramic datasets remain relatively scarce, often orders of magnitude smaller: PASS has only 1,050 annotated panoramas for segmentation, WildPASS 2,500 samples, and Deep360 merely 2k panoramic depth maps. 
Even in generation and super-resolution tasks, panoramic datasets usually contain only a few thousand samples, compared with hundreds of thousands or even millions on the perspective side.

This imbalance highlights a critical bottleneck: the lack of large-scale, diverse, and richly annotated panoramic datasets hinders the development of generalizable models and fair benchmarking across tasks. 
While perspective vision has benefited greatly from the scale of data, panoramic vision is still constrained by limited data availability, preventing models from fully exploiting the potential of 360\textdegree{} understanding and generation. 
Bridging this gap is therefore an urgent priority for the community, calling for systematic efforts in panoramic data collection, annotation, and benchmark design.

\section*{Future Work by Task}

\subsection*{Visual Quality Enhancement and Assessment}

As summarized in Fig. \ref{fig:future_1}, future research on visual quality enhancement and assessment for panoramic vision can be explored along several directions.

\noindent \textbf{Super-Resolution:} Leverage diffusion models with textual guidance for semantically aware panoramic restoration, improve real-time efficiency via pruning and architectural optimization, and incorporate modules specialized for key categories such as human faces.  

\noindent \textbf{Reflection Removal:} Move toward reflection separation by leveraging text or auxiliary cues to disentangle reflection and transmission, while extending to outdoor scenarios with semantic and multimodal understanding to improve robustness and generalization.  

\noindent \textbf{Image Restoration:} Address diverse degradations (e.g., rain, snow, fog, low-light). Current models are often single-task and trained on synthetic data, highlighting the need for more generalized all-in-one approaches and unsupervised or self-supervised frameworks that adapt to real-world panoramic conditions.  

\noindent \textbf{Visual Quality Assessment:} FR-OIQA often assumes perfect references, which is unrealistic given sensor noise and stitching artifacts; future work should jointly assess fidelity and perceptual naturalness, particularly in generative enhancement scenarios. Meanwhile, NR-OIQA suffers from limited supervision and weak generalization; integrating vision–language pipelines with LLMs can provide semantic priors and subjective cues to improve distortion reasoning and quality prediction.  

\begin{figure}[t]
    \centering
    \includegraphics[width=\linewidth, keepaspectratio]{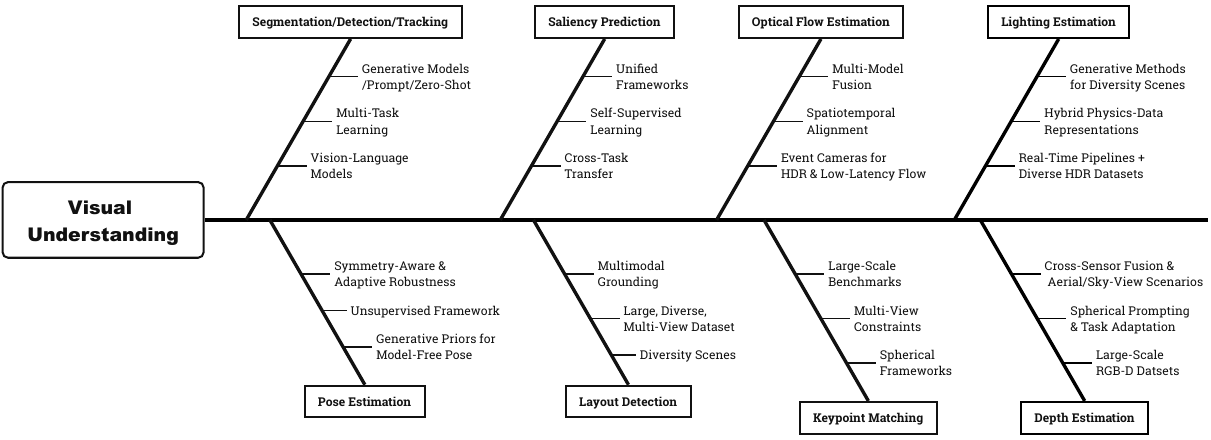}
        \vspace{-15pt}
    \caption{Summary of Future Directions in Visual Understanding.}
    \label{fig:future_2}
    \vspace{-15pt}
\end{figure}

\subsection*{Visual Understanding}

Future directions in panoramic visual understanding are summarized in Fig. \ref{fig:future_2}.

\noindent \textbf{Segmentation:} Use generative models (e.g., diffusion, masked autoencoders) to improve representation learning and zero-shot transfer; adopt multi-task learning with generation tasks (e.g., inpainting, novel view synthesis) to yield richer features; and leverage large vision–language models for open-world segmentation with weakly supervised multimodal alignment.  

\noindent \textbf{Detection:} Employ prompt-driven detection for flexible zero-shot queries; integrate multimodal inputs (RGB, depth, thermal, text) to improve robustness and generalization; and adopt open-vocabulary detection to foster lifelong learning and uncertainty awareness. Future progress hinges on integrating these capabilities with large-scale vision–language understanding and continual learning for safe, scalable detection.  

\noindent \textbf{Tracking:} Explore zero-shot paradigms with vision–language models for prompt-based initialization and semantic reasoning, unified frameworks integrating detection, tracking, and segmentation for spatiotemporal consistency, and cross-view or multimodal fusion for robustness under occlusion. Uncertainty-aware designs further enable long-term, real-world deployment, pointing toward more open, semantic-aware, and generalizable tracking systems.  

\noindent \textbf{Pose Estimation:} Improve generalization and reduce annotation reliance by leveraging generative priors for model-free object pose, enhance robustness in category-level tasks with symmetry-aware and adaptive methods, and advance unsupervised facial/head pose estimation via geometric cues, temporal coherence, and pre-trained representations.  

\noindent \textbf{Saliency Prediction:} Develop unified frameworks combining distortion-aware architectures, dynamic attention, and multimodal cues under spherical representations, while leveraging self-supervised panoramic video learning and cross-task transfer (e.g., saliency to scanpath) for robust and human-aligned saliency modeling.  

\noindent \textbf{Layout Detection:} Generalize beyond Manhattan-aligned scenes to non-Manhattan, multi-level, or dynamic layouts. Dataset scale and diversity remain insufficient, highlighting the need for larger, richly annotated multi-view panoramas. Integrating multimodal cues (e.g., depth, audio, inertial data, language) could further ground layout estimation in embodied perception, enabling more intelligent agents in immersive environments.  

\begin{figure}[t]
    \centering
    \includegraphics[width=0.9\linewidth, keepaspectratio]{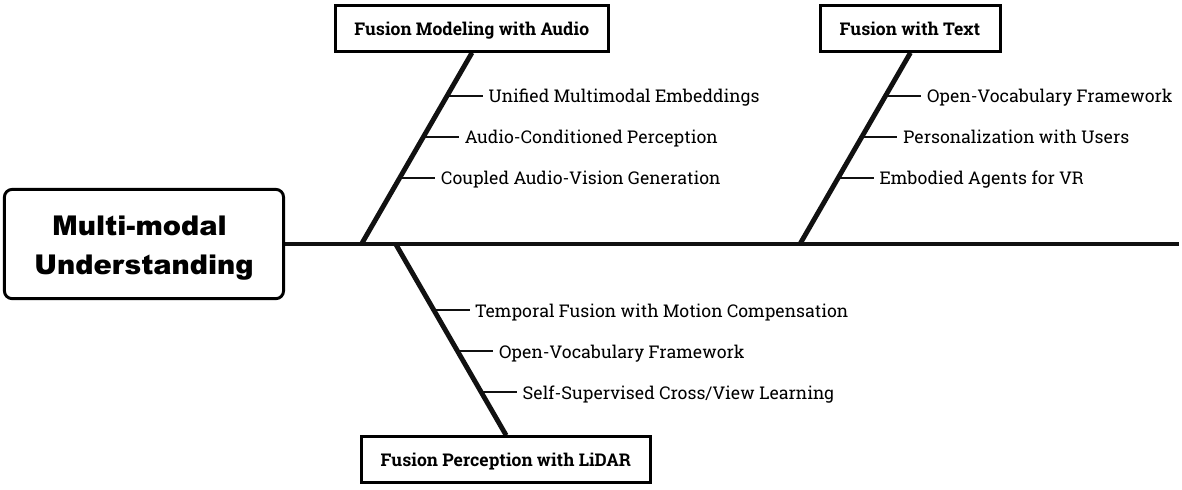}
        \vspace{-5pt}
    \caption{Summary of Future Directions in Multi-modal Understanding.}
    \label{fig:future_3}
    \vspace{-15pt}
\end{figure}

\noindent \textbf{Optical Flow Estimation:} Advance multi-modal panoramic flow by integrating RGB-D cues for robustness under occlusion or low light, and leverage event cameras for low-latency, HDR motion capture. Networks that fuse event streams with image-based representations via spatiotemporal alignment or neural fusion, supported by dedicated RGB-Event or RGB-D datasets, could enable more consistent and reliable flow estimation in challenging conditions.  

\noindent \textbf{Keypoint Matching:} Develop spherical frameworks with rotation-equivariant features, incorporate multi-view constraints for robust correspondence, and build large-scale benchmarks with ground-truth panoramic matches.  

\noindent \textbf{Decomposition:} Develop unified spherical models capable of handling dynamic scenes and complex materials, supported by larger and more diverse panoramic datasets.  

\noindent \textbf{Lighting Estimation:} Current generative methods remain limited under diverse outdoor and dynamic lighting, motivating the use of temporal cues and multimodal signals. Physically inspired approaches need hybrid physics–data models for greater realism and interpretability. Structured pipelines should be optimized for real-time, lightweight VR/AR deployment, while diverse HDR panoramic datasets with scene–illumination pairs are essential for robust generalization.  

\noindent \textbf{Depth Estimation:} Tackle the shortage of high-quality RGB-D panoramas by building large-scale datasets for foundation models. Advances may come from spherical prompting and task adaptation for better transfer, cross-sensor fusion with LiDAR or IMU for real-world robustness, and extending beyond ground-level to aerial and sky-view scenarios.

\begin{figure}[t]
    \centering
    \includegraphics[width=\linewidth, keepaspectratio]{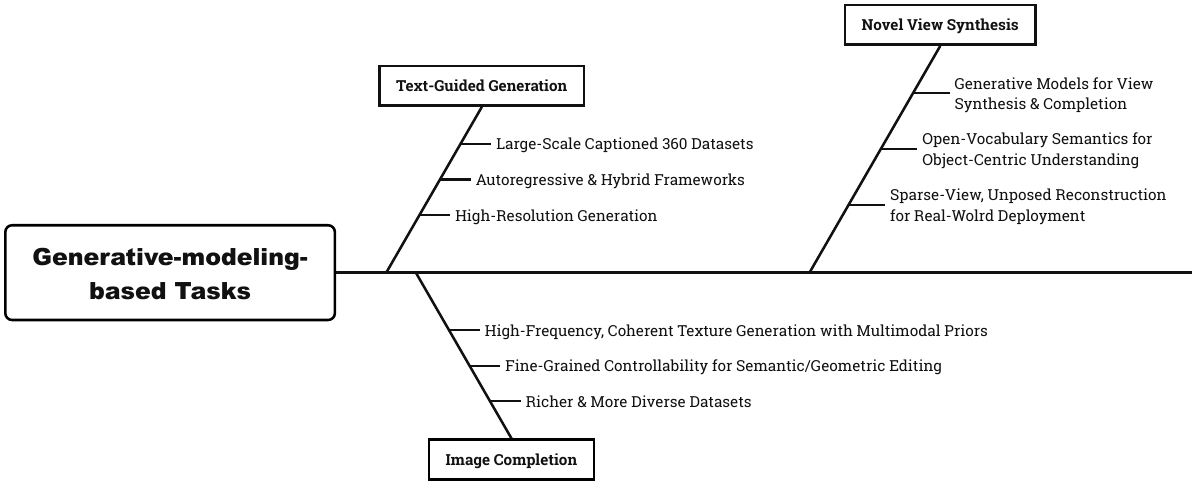}
        \vspace{-15pt}
    \caption{Summary of Future Directions in Generative-modeling-based Tasks.}
    \label{fig:future_4}
    \vspace{-10pt}
\end{figure}

\subsection*{Multi-modal Understanding} 

As illustrated in Fig. \ref{fig:future_3}, multi-modal understanding provides rich future directions.

\noindent \textbf{Fusion Modeling with Audio:} Pursue unified embeddings of spatial, semantic, and temporal cues, audio-conditioned perception for attention and control, and real-time multimodal modeling for embodied AI. Key challenges include reverberation, occlusion-aware propagation, and adaptive feedback, while coupled audio–vision generation (e.g., text-conditioned 4D synthesis with ambient sound) offers another promising path.  

\noindent \textbf{Fusion Perception with LiDAR:} Explore self-supervised learning with cross-view/modal consistency for robust features, open-vocabulary detection and segmentation via vision–language integration, and temporal fusion with motion compensation to improve dynamic scene perception.  

\noindent \textbf{Fusion with Text:} Advance open-vocabulary detection and viewpoint-aware grounding, develop lightweight and distortion-aware models for efficient deployment, enhance personalization through user intent and preference alignment, and enable embodied agents for VR instruction following, spatial dialogue, and multimodal 360\textdegree{} interaction.

\subsection*{Generative-modeling-based Tasks}

Finally, as summarized in Fig. \ref{fig:future_4}, generative modeling for panoramic data opens several new avenues.

\noindent \textbf{Text-guided Generation:} Build large-scale, captioned 360\textdegree{} datasets to improve scalability and semantic grounding; explore autoregressive and hybrid frameworks beyond diffusion for long-range spherical modeling; and advance high-resolution generation by overcoming memory and architectural constraints. Tackling these directions will enable semantically aligned, visually consistent panoramic content for diverse downstream applications.  

\noindent \textbf{Image Completion:} Expand dataset scale, diversity, and realism to improve generalization; enhance controllability for fine-grained semantic and geometric editing; and improve the generation of high-frequency, coherent textures over large missing regions. Promising directions include building richer datasets, developing interactive and viewpoint-aware frameworks, and integrating external priors with multimodal cues for more realistic and controllable 360\textdegree{} scene completion.  

\noindent \textbf{Novel View Synthesis:} Leverage generative models for view synthesis and completion, integrate open-vocabulary semantics for object-centric understanding, jointly model motion and geometry for dynamic scenes, and reduce reliance on dense posed inputs to enable sparse-view, unposed reconstruction in real-world settings.

\end{document}